\documentclass[acmsmall,screen,nonacm]{acmart}
\usepackage{graphicx}
\usepackage{natbib}
\usepackage{amsmath}
\usepackage{amsfonts}
\usepackage{booktabs}
\usepackage{multirow}
\usepackage{makecell}
\usepackage{url}
\usepackage{diagbox}
\usepackage{tabu}
\usepackage{xcolor}
\usepackage[ruled,vlined,noend]{algorithm2e}
\newcommand{\Blue}[1]{{\color{blue}#1}}

\newcommand{\Black}[1]{{\color{black}#1}}
\newcommand{\Green}[1]{{\color{green}#1}}
\definecolor{applegreen}{rgb}{0.55, 0.71, 0.0}
\definecolor{green}{rgb}{0.0, 0.5, 0.0}

\begin{document}

\title{Triples-to-Text Generation with Reinforcement Learning Based Graph-augmented Structural Neural Networks}

\author{Hanning Gao}
\authornote{Both authors contributed equally to this research.}
\email{ghnqwerty@gmail.com}
\orcid{0000-0002-6621-4998}
\affiliation{%
   \department{Department of Computer Science and Technology, College of Electronic and Information Engineering}
  \institution{Tongji University}
  \streetaddress{4800 Cao'an Road}
  \city{Shanghai}
  \country{China}
  \postcode{201804}
}

\author{Lingfei Wu}
\authornotemark[1]
\email{lwu@email.wm.edu}
\affiliation{%
  \institution{JD.COM Silicon Valley Research Center}
  \streetaddress{675 E Middlefield Rd}
  \city{Mountain View}
  \state{CA}
  \country{USA}
  \postcode{94043}}
\email{lwu@email.wm.edu}

\author{Hongyun Zhang}
\authornotemark[2]
\email{zhanghongyun@tongji.edu.cn}
\affiliation{%
  \department{Department of Computer Science and Technology, College of Electronic and Information Engineering}
  \institution{Tongji University}
  \streetaddress{4800 Cao'an Road}
  \city{Shanghai}
  \country{China}
  \postcode{201804}
}

\author{Zhihua Wei}
\authornotemark[2]
\email{zhihua\_wei@tongji.edu.cn}
\affiliation{%
  \department{Department of Computer Science and Technology, College of Electronic and Information Engineering}
  \institution{Tongji University}
  \streetaddress{4800 Cao'an Road}
  \city{Shanghai}
  \country{China}
  \postcode{201804}
}

\author{Po Hu}
\email{phu@mail.ccnu.edu.cn}
\affiliation{%
  \department{Hubei Provincial Key Laboratory of Artificial Intelligence and Smart Learning, Central China Normal University}
  \department{School of Computer}
  \institution{Central China Normal University}
  \city{Wuhan}
  \state{Hubei}
  \country{China}
}

\author{Fangli Xu}
\email{lili@yixue.us}
\affiliation{%
 \institution{Squirrel AI Learning}
 \city{NY}
 \country{USA}}

\author{Bo Long}
\email{bo.long@jd.com}
\affiliation{%
  \institution{JD.COM}
  \city{Beijing}
  \country{China}}


\begin{abstract}
Considering a collection of RDF triples, the RDF-to-text generation task aims to generate a text description.
Most previous methods solve this task using a sequence-to-sequence model or using a graph-based model to encode RDF triples and to generate a text sequence. Nevertheless, these approaches fail to clearly model the local and global structural information between and within RDF triples. 
Moreover, the previous methods also face the non-negligible problem of low faithfulness of the generated text, which seriously affects the overall performance of these models.
To solve these problems, we propose a model combining two new graph-augmented structural neural encoders to jointly learn both local and global structural information in the input RDF triples.
To further improve text faithfulness, we innovatively introduce a reinforcement learning (RL) reward based on information extraction (IE).
We first extract triples from the generated text using a pretrained IE model and regard the correct number of the extracted triples as the additional RL reward.
Experimental results on two benchmark datasets demonstrate that our proposed model outperforms the state-of-the-art baselines,  and the additional reinforcement learning reward does help to improve the faithfulness of the generated text.
\end{abstract}




\keywords{RDF-to-text generation, data-to-text generation, graph neural networks (GNN), graph to sequence (Graph2Seq), reinforcement learning (RL), text faithfulness.}

\maketitle
\pagestyle{plain}
\section{Introduction}

As a sub-task of data-to-text generation, RDF-to-text generation transforms a collection of Resource Description Framework (RDF) triples into an faithful and informative text. RDF triple is a popular representation of knowledge graph, with the form of (a subject entity, a relationship, an object entity). Therefore, a given collection of RDF triples can naturally form a graph, thus allowing RDF-to-text generation to evolve into a graph-to-sequence generation problem.

\begin{figure}[ht]
\centering
\includegraphics[scale=0.3]{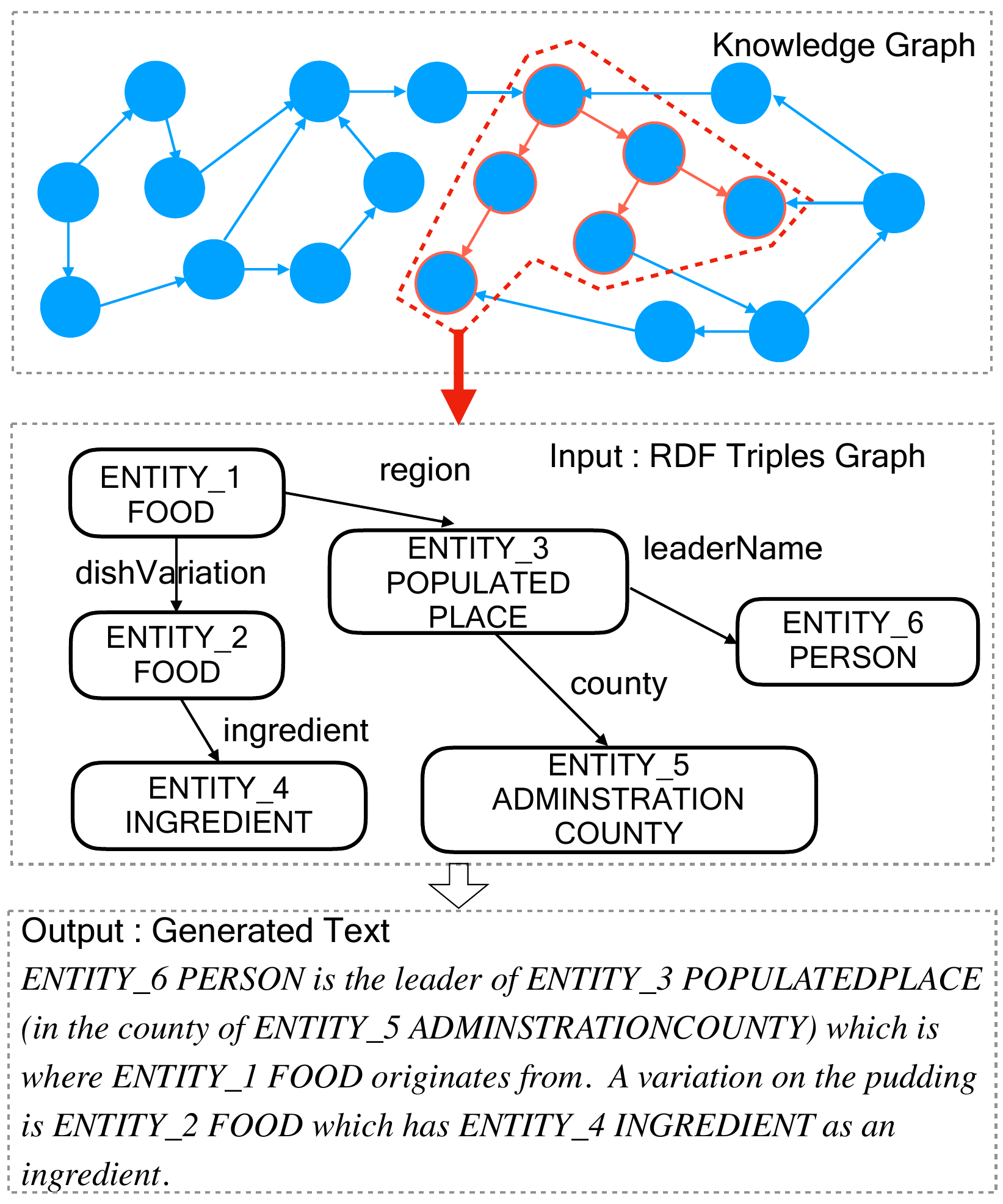}
\caption{An RDF triples graph formulated by a collection of RDF triples extracted from a large-scale knowledge graph and its corresponding text descriptions. Entity masking (introduced in Sec.\ref{Entity-Masking}) has been applied in this example.}
\label{fig:examplegraph}
\end{figure}

RDF-to-text generation is challenging task because both the graph structure and the semantic information of RDF triples needs to be well modeled. The example in Figure \ref{fig:examplegraph} shows an RDF triples graph extracted from a large-scale knowledge graph with its text description, where the nodes (e.g. ``\textit{ENTITY\_1 FOOD}" and ``\textit{ENTITY\_4 INGREDIENT}") represent entities with their corresponding entity types and the edges (e.g. ``\textit{dishVariation}" and ``\textit{ingredient}") represent the relationships connecting pairs of entities. This example is centered on the ``\textit{ENTITY\_1 FOOD}'' entity node, which is referred to as the topic entity for this example. In recent years, RDF-to-text generation has attracted increasing attention because of its fact-aware applications in knowledge-based question answering~\cite{hao2017end}, data-driven text generation~\cite{liu2018table} and entity summarization~\cite{pouriyeh2017lda}.

RDF-to-text generation can be solved using the classical text generation paradigm, which primarily concentrate on content selection \cite{duboue2003statistical} and surface realization \cite{deemter2005real}. However, the critical issue of error propagation has been largely ignored, to the harm of the quality of generated texts. With the recent tremendous advances in end-to-end deep learning in the field of natural language processing, promising results have been achieved in RDF-to-text generation through the use of various sequence-to-sequence (Seq2Seq) models \cite{gardent2017webnlg, jagfeld2018sequence, ferreira2019neural}. In order to feed into Seq2Seq models, the individual RDF triples in the collection need to be connected end-to-end into a sequence.

However, simply converting the collection of RDF triples into a sequence may result in the loss of important structural information. As a collection of RDF triples can be formed as a input graph, a number of graph-based methods have recently been proposed to integrate structural information for this task. \citet{trisedya2018gtr} proposed a triple encoder, GTR-LSTM, which captures entity relationships within and between triples by sampling different meta-paths, thus retaining the graph structure in the graph-based encoder.
In addition, because the component of GTR-LSTM is still based on recurrent neural networks (RNNs), it is often unable to model the local complex structural information within triples.

On the other side, to directly encodes the collection of RDF triples of graph structure and decodes a text sequence, DGCN presented in \cite{marcheggiani2018deep} is a graph-to-sequence (Graph2Seq) model based on an improved Graph Convolutional Networks (GCN) ~\cite{kipf2016semi}. However, it is well-known that when using multi-layers ($>$=3 layers), GCN often tends to over-fit quickly and its ability of learning longer range dependency is weakened as well. As a result, this GCN model generally performs better at modeling local information of the graph formed by triples than it does at modeling global information between RDF triples. 

{Although great efforts have been made to build effective models to capture both structural and semantic information in the input triples, generating an informative and faithful text is still a challenge.
Through manual analysis of the generated texts, we find that the mistakes made by these end-to-end neural network models can be divided into two categories: (1) \textit{Misattribution.} The subject or object corresponding to the relationship in the generated text is mistaken. For example, given a triple ``\textit{(ENTITY\_1 POLITICIAN, birthPlace, ENTITY\_2 BIRTHPLACE)}'', the generated text is \textit{ENTITY\_1 is the birthplace of ENTITY\_1}, which obviously misidentifies the object entity.
(2) \textit{Information missing.} The model fails to cover the salient information of all input triples with the number of triples increasing.}

In this study, we present a reinforcement learning based graph-augmented structural neural encoders framework for RDF-to-text generation to address the aforementioned issues. We first propose to harness the power of graph-based meta-paths encoder and graph convolutional encoder to jointly model both local and global structural information. Then, separated attentions are applied to two different encoder outputs to learn the final attentional hidden representation at each decoding time step. 
Moreover, we also design a selection mechanism inspired by pointer-generator \cite{see2017get}, which controls the degree of attention given to graphs or meta-paths at each step of decoding.
{To improve the faithfulness of the generated text, we innovatively propose an information extraction (IE) based reinforcement learning reward. We first pretrain an IE model with ground-truth texts as inputs and triples as their predicted outputs on the same datasets. Then, we use the number of correct triples predicted by the IE model as the RL reward. Finally, we train the graph-augmented structural neural encoders by optimizing a hybrid objective function that combines both RL loss and cross entropy loss.}

In this paper, our proposed graph-augmented structural neural encoders can concentrate on two different perspectives of the collection of RDF triples. A novel bidirectional Graph-based Meta-Paths (bi-GMP) encoder is used to capture global long-range dependency between the input triples, while a new bidirectional Graph Convolutional Network (bi-GCN) encoder mainly concentrates on explicitly modeling the local structural information within the triples. 
{We further exploit an IE-based RL reward to address the two types of faithfulness mistakes summarised in the previous paragraph to a certain extent.}

The main contributions we highlight are listed below:
\begin{itemize}
\item We present a {RL-based} graph-augmented structural neural encoders framework by combining a bi-GMP encoder and a bi-GCN encoder for explicitly capturing the global and local structural information in the input RDF triples.  

\item Separated attention mechanism is applied on the graph encoders and their corresponding context vectors are fused using a selection mechanism to better decode the text descriptions. 

\item {We propose an IE-based RL reward by computing the correct number of triples in the generated texts and explore the effect of this reward on RDF triple sets of different sizes.}

\item  The experimental results on two RDF-to-text datasets WebNLG and DART demonstrate the advantages of our model in BLEU, METEOR and TER metrics and faithfulness metric. 

\end{itemize}

\section{Related Work}
Our method is closely related with researches in the fields of RDF-to-text generation, graph neural networks (GNN) for text generation and information extraction.

\subsection{RDF-to-Text Generation}
The aim of RDF-to-text generation is to generate a informative, grammatically correct, faithful and fluent description for a given collection of RDF triples. Web Ontology Language to text generation \cite{bontcheva2004automatic, stevens2011automating, androutsopoulos2013generating} and knowledge base verbalization~\cite{banik2012kbgen} were early forms of this task. Pipeline system is often used to address data-to-text generation, including two main steps: (1) \textit{content selection} addresses the question of what content should be described \cite{barzilay2005collective}; (2) \textit{surface realization} realizes the text generation process token by token \cite{deemter2005real, belz2008automatic}. 

Over the past few years, a number of natural language generation (NLG) tasks based on Seq2Seq model with attention mechanism \cite{bahdanau2014neural, luong2015effective} and copy mechanism \cite{see2017get, gu2016incorporating} have achieved promising performance. A range of research work such as \cite{gardent2017webnlg, jagfeld2018sequence} show that Seq2Seq model and its variant models perform prominently on RDF-to-text generation by flattening a collection of RDF triples into a text sequence. \citet{ferreira2019neural} implemented two encoder-decoder architectures using Gated-Recurrent Units (GRU) \cite{cho-etal-2014-properties} and Transformer \cite{vaswani2017attention} respectively. To improve the probability of producing high-quality texts, \citet{zhu2019triple} proposed a model minimizing the Kullback-Leibler (KL) divergence between the distributions of the real text and generated text. \citet{moryossef2019step} proposed a model combining the pipeline system and neural networks to match a reference text with its corresponding text plan to train a plan-to-text generator. 
{
In addition, some work has focused solely on improving the performance of the seen split part of dataset, where the entity types have appeared in the training set.
\citet{ribeiro2020modeling} combined two GNN encoders to encode both global and local node contexts for WebNLG seen categories.
}
{Recently, data-to-text generation has benefited from large-scale pre-trained language models (PLMs). \citet{wang2021stage} and \citet{kale2020text} applied T5 \cite{raffel2019exploring} in this task. Moreover, \citet{wang2021stage} further exploited external knowledge from Wikipedia to improve the model performance. }


\subsection{Graph Neural Networks}
There has been a significant increase in interest in using graph neural networks for data-to-text generation. 

\textbf{Graph Neural Networks.} Graph Neural Networks have achieved excellent performance on many natural language processing tasks where the input data can be represented as a graph. Essentially, graph neural networks are graph representation learning models that learn each node embedding using graph structure and its neighbouring node embeddings. According to \cite{ma2021deep}, the learning process of node embeddings is referred as \textit{graph filtering}, which can refine the node embeddings without modifying the graph structure. Graph filters can be roughly categorized into four classes~\cite{wu2021graph}: spectral-based like GCN \cite{kipf2016semi}, spatial-based like GraphSage \cite{hamilton2017inductive}, attention-based like Graph Attention Network (GAT) \cite{velivckovic2017graph} and recurrent-based like Gated Graph Neural Network (GGNN) \cite{li2015gated}. Furthermore, \textit{graph pooling} operation aggregates node embeddings and produces a smaller graph with fewer nodes inspired by CNNs \cite{krizhevsky2012imagenet}. When the smaller graph only contains one node, the node embedding is considered as the graph-level embedding for the input graph.

\textbf{Graph to Sequence Learning.}
Sequence-to-Sequence model is a widely used encoder-decoder framework in the natural language generation field. However, when the input is graph data such as dependency graph, Abstract Meaning Representation (AMR) \cite{banarescu-etal-2013-abstract} graph and knowledge graph, it often needs to be linearized into a sequence before it can be fed into the Seq2Seq model, resulting in a loss of structural information in the graph. To alleviate the limitation of Seq2Seq models on encoding graph data, a number of graph-to-sequence (Graph2Seq) models have been proposed \cite{bastings2017graph, beck2018graph, song2019semantic, yao2020heterogeneous}.
\citet{bastings2017graph} applied GCN on dependency trees of input sentences to produce word representations.
\citet{beck2018graph} proposed a graph-to-sequence model based on GGNN with edges being turned into additional nodes. 
\citet{song2019semantic} explored the effectiveness of AMR as a semantic representation for neural machine translation based on a graph recurrent network.
Previous studies \cite{chen2019reinforcement, gao2019dyngraph2seq,xu2018graph2seq} applied a bidirectional graph encoder to encode an input graph and then used an attention-based LSTM \cite{hochreiter1997long} decoder. 
To capture longer-range dependencies, \citet{song2018graph} employed an LSTM to process the state transitions of the graph encoder outputs. \citet{marcheggiani2018deep} used a relational graph convolutional network, which is introduced and extended in~\cite{kipf2016semi, bruna2013spectral,defferrard2016convolutional}, to encode each node using its neighboring nodes, edge directions and edge labels simultaneously.

\subsection{Information Extraction}
\label{related work IE}
The aim of Information Extraction (IE) is to extract entity pairs and their relationships within a given sentence. Information extraction is a significant task in the field of natural language processing, as it helps to mine factual knowledge from free texts. Information extraction is an inverse task of RDF-to-text generation, and there are a number of researches on both tasks on the same dataset.

As two important subtasks of information extraction, named entity recognition (NER) and relation extraction (RE) can be operated via a pipeline approach that firstly recognize entities and then conduct relation extraction \cite{zelenko2003kernel, chan2011exploiting}. NER predicts a label for each word in a sentence indicating whether it is an entity or not. For a sentence with annotated entities, RE is essentially a relation classification task which predicts a relation label for each entity pairs. \citet{zeng2014relation}, \citet{xu2015classifying} and \citet{sahu2019inter} conducted relation extraction via CNN, RNN and GNN, respectively.

Recently, researchers start to explore joint entity recognition and relation extraction. 
Some Seq2Seq-based methods \cite{zeng2018extracting, nayak2020effective, zeng2020copymtl, sui2020joint} directly generate relational triples. For example, Set Prediction Networks (SPN) \cite{sui2020joint} is featured by transformers, which can directly predict the final collection of triples. 
Other models are tagging based methods \cite{zheng2017joint, dai2019joint, wang2020tplinker}, which regard the IE task as a sequence labeling problem and use specially designed tagging schema. For example, TPLinker \cite{wang2020tplinker} is a one-stage joint extraction model with a hand-crafted tagging schema and can discover overlapping relations sharing one or both entities. There are also other methods that represent entities and relationships in a shared parameter space, but extract the entities and relationships respectively \cite{miwa2016end, zhang2017end, zhao2021representation}. For example, \citet{zhao2021representation} proposed a representation iterative fusion strategy on heterogeneous graph neural networks for relation extraction. After obtaining the final representations of entity nodes and relation nodes,
this model uses specialised taggers to extract subjects and objects, respectively. In this paper, we pretrain the SPN on the same dataset of the RDF-to-text generation task to extract triples from the generated texts.

\section{Problem Formulation}
Next, we give a formal definition of the RDF-to-text generation task. The input is a collection of RDF triples denoted as $S=\{t_1,t_2,...t_n\}$, where $t_i=(s_i, r_i, o_i)$ is a triple containing a subject entity $s_i$, a relationship $r_i$ and an object entity $o_i$. A collection of RDF triples is represented as a directed graph. The aim of this task is to generate a text description $\hat{Y}=\left \langle w_1,w_2,...w_T \right \rangle$ which maximizes the conditional likelihood:
\begin{equation} \label{eq1}
    \hat{Y}=\text{argmax}_Y P(Y|S)
\end{equation}
The genereted text $\hat{Y}$ should represent the comprehensive and correct information of entities and relationships in the input collection of RDF triples. 

In this paper, we construct two input graphs based on a collection of RDF triples. The bi-GMP encoder takes input graph $\mathcal{G}_1 = (\mathcal{V}_1, \mathcal{E}_1)$ as input, where $\mathcal{V}_1$ denotes the entity nodes set and $\mathcal{E}_1$ represents original relationships. Whilst, the bi-GCN encoder takes input graph $\mathcal{G}_2 = (\mathcal{V}_2, \mathcal{E}_2)$ as its input, where $\mathcal{V}_2$ is composed of entity nodes and relationship nodes because relationships are seen as newly added nodes, not as edges. $\mathcal{E}_2$ indicates a predefined set of edges that describes the relationship between an entity node and a relationship node, or the relationship between multiple tokens of a node. More details about the construction of the input graphs are depicted in Sec.\ref{s:Graph Constructions}. Therefore, Equation \ref{eq1} can also be written as follows:
\begin{equation} \label{eq2}
    \hat{Y}=\text{argmax}_Y P(Y|\mathcal{G}_1, \mathcal{G}_2)
\end{equation}

{Furthermore, we give the notation of information extraction task, whose goal is to extract all possible triples in a given sentence. Considering an input sentence $Y$, the IE model aims to extract the target RDF triple set $\hat{S}=\{\widehat{t_1},\widehat{t_2},...\widehat{t_n}\}$. }

\section{Our Proposed Model}
\begin{figure*}
\centering
\includegraphics[scale=0.26]{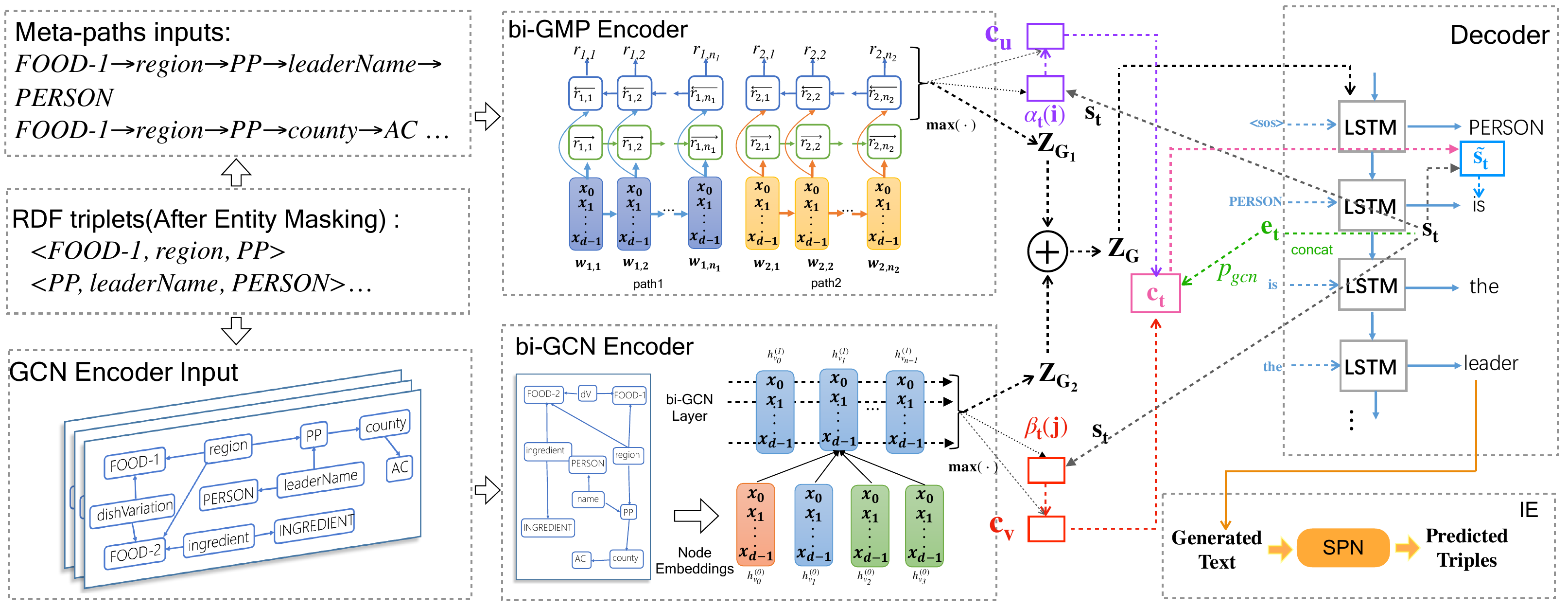}
\caption{The framework of the combined Graph-augmented Structural Neural Encoders Model. This framework contains four key components: a graph construction module, a combined graph-augmented encoders module, a decoder module and a post information extraction module.} 
\label{fig:combined_modle}
\end{figure*}

\subsection{Architecture of Our Proposed Model}
As shown in Fig.\ref{fig:combined_modle}, our proposed model contains four key components: a graph construction module, a combined graph-augmented encoders module, a decoder module and a post information extraction module. First, the graph construction module leverages the input RDF triples to build two input graphs. Second, after graph construction, we propose two graph-based encoders to model the input graphs and learn node embeddings and graph embeddings. Third, we employ an LSTM-based decoder to generate the text word by word. In this module, we propose separated attention mechanism on both graph-based encoders and fuse their context vectors using a selection mechanism. Fourth, a pretrained information extraction model, i.e.SPN extracts triples from the sequence generated by sampling and the sequence generated by greedy search. A reinforcement learning reward is computed based on the correct numbers of extracted triples.

In the next few sections, we first propose a graph-augmented structural neural encoders framework including two graph-based encoders. Then, we describe how to combine the two encoders for better modeling global and local information. Finally, we introduce how to compute the reward of the whole model using the pretrained information extraction model.

\subsection{Bidirectional Graph-based Meta-Paths Encoder}
Given a collection of RDF triples, it can be formed as a directed graph naturally with subjects and objects being nodes and relationships being edges. We refer this naturally formed graph as $\mathcal{G}_1$. For the graph $\mathcal{G}_1=(\mathcal{V}_1,\mathcal{E}_1)$, $\mathcal{V}_1$ represents the entity nodes set, and $\mathcal{E}_1$ represents original relationships between entities.
With the purpose of encoding information depending on different meta-paths and model long-range dependencies in $\mathcal{G}_1$, we propose a novel bidirectional Graph-based Meta-Paths encoder (bi-GMP) enhanced on the basis GTR-LSTM \cite{trisedya2018gtr}. In contrast to GTR-LSTM, the bi-GMP encoder performs hidden state masking on different meta-paths to ensure the encoding of different meta-paths do not interfere with each other.

\begin{algorithm}
\caption{Meta-paths Computing Algorithm} \label{alg:Meta-paths}
\SetAlgoLined
\textbf{Input:} Input graph $\mathcal{G}_1=(\mathcal{V}_1,\mathcal{E}_1)$ \\
Set empty node set $\mathcal{V}_{\text{IN}}$ for zero in-degree nodes; \\
Set empty node set $\mathcal{V}_{\text{OUT}}$ for zero out-degree nodes; \\
\For{node ${v}$ in $\mathcal{V}$}{
Compute the in-degree $d_i$ and out-degree $d_o$ for node $v$; \\
\If{$d_i$ is equal to 0}{Add $v$ to $\mathcal{V}_{\text{IN}}$; \\}
\If{$d_o$ is equal to 0}{Add $v$ to $\mathcal{V}_{\text{OUT}}$; \\}
}
Set meta-path set $S_p = \{\}$; \\
\For{node $v_i$ in $\mathcal{V}_{\text{IN}}$}{
\For{node $v_j$ in $\mathcal{V}_{\text{OUT}}$}
{Find the shortest path $p_k$ from $v_i$ to $v_j$; \\
Add $p_k$ to $S_p$; \\
}
}
\end{algorithm}

We select the meta-paths of $\mathcal{G}_1$ using a combination of topological sort algorithm and single-source shortest path algorithm shown in Algo.\ref{alg:Meta-paths}. Firstly, we find a node set $\mathcal{V}_{\text{IN}}$ with zero in-degree nodes and a node set $\mathcal{V}_{\text{OUT}}$ with zero out-degree nodes. Secondly, by choosing a starting node from $\mathcal{V}_{\text{IN}}$ and a target node from $\mathcal{V}_{\text{OUT}}$, we calculate a single-source shortest path with both nodes and relationships retained in the path. Finally, a sequence containing a set of meta-paths $S_p = \{p_1, p_2, ..., p_k, ...\}$ can be selected from the graph $\mathcal{G}_1$, where $p_k=\left \langle w_{k, 1},w_{k, 2},...w_{k, n_k}\right \rangle$. {The hidden state masking means that the hidden state of the final token in $p_{k-1}$ is masked when to compute the hidden state of the first token in $p_k$.}

After obtaining the meta-paths, we initialize the word embeddings using 6B GloVe vectors\footnote{\url{https://nlp.stanford.edu/projects/glove/}} for each word in $S_p$. Each word $w_{k,i}$ is represented as a distributed representation $\mathbf{w}_{k,i} \in \mathbb{R}^{d_e}$, where the dimension $d_e$ is equal to 300. Taking inspiration from the benefits of BiLSTM over LSTM, the meta-paths encoder computes the token representations in both directions for each meta-path and concatenates them together in the last step. 
When the word embedding $\mathbf{w}_{k,i}$ is given, the hidden state $\mathbf{r}_i$ is computed as follows:
\begin{align}
    \overrightarrow{\mathbf{r}_{k, i}} &= f(\overrightarrow{\mathbf{r}_{k, i-1}}, \mathbf{w}_{k, i}) \label{eq3}\\
    \overleftarrow{\mathbf{r}_{k, i}} &= g(\overleftarrow{\mathbf{r}_{k, i+1}}, \mathbf{w}_{k, i}) \label{eq4}\\
    \mathbf{r}_i &= \text{concat}(\overrightarrow{\mathbf{r}_{k, i}}, \overleftarrow{\mathbf{r}_{k, i}})
\end{align}
where 
$f(\cdot)$ and $g(\cdot)$ are single LSTM units and $\text{concat}(\cdot)$ is the concatenation operation. If the previous or the next hidden state belongs to different meta-path, an all-zero hidden state $\mathbf{r}_0$ is input to $f(\cdot)$ or $g(\cdot)$. Equation \ref{eq3} and Equation \ref{eq4} can be written as:
\begin{align}
    \overrightarrow{\mathbf{r}_{k, 1}} &= f(\mathbf{r}_{0}, \mathbf{w}_{k, 1})\\
    \overleftarrow{\mathbf{r}_{k, n_k}} &= g(\mathbf{r}_{0}, \mathbf{w}_{k, n_k})
\end{align}
where $n_k$ is the length of meta-path $p_k$.

Finally, a set of entity node representations $R_{1} = [\mathbf{r}_1; \mathbf{r}_2; ...; \mathbf{r}_{L_1}]^T \in \mathbb{R}^{L_1 \times D}$ is output by the bi-GMP encoder, where $L_1$ is the expanded length of $S_p$. The graph embedding of graph $\mathcal{G}_1$ is computed over $R_{1}$:
\begin{equation}
    Z_{\mathcal{G}_1} = \text{maxpool}(R_1)
\end{equation}
where $\text{maxpool}(\cdot)$ is the max pooling operation over the first dimension and $Z_{\mathcal{G}_1} \in \mathbb{R}^D$.

\subsection{Bidirectional Graph Convolutional Networks Encoder}
To better use the information in relationships, we build the input graph for bi-GCN encoder by converting relationships to nodes in the graph. For the graph $\mathcal{G}_2=(\mathcal{V}_2, \mathcal{E}_2)$, relationships are considered as graph nodes. Therefore, $\mathcal{G}_2$ is composed of entity nodes and relationship nodes. Edges in $\mathcal{E}_2$ may connect relationship node to entity node or entity node to entity node.
More details about the construction of bi-GCN graph are illustrated in Sec.\ref{RDF Triples to bi-GCN Graph}.

To effectively learn the node embeddings and the graph embeddings for the constructed graph, we present the bi-GCN encoder based on the conventional GCN, which learns node embeddings from both incoming and outgoing edges. At each layer, we first compute the intermediate node representations $\mathbf{H}_{\vdash}^{(l)}$ and $\mathbf{H}_{\dashv}^{(l)}$ in both directions. Then, we fuse the two node representations at this layer to obtain the fused node representations $\mathbf{H}^{(l)}$. In particular, the fused node representations at layer $l$ can be represented as ${\mathbf{H}^{(l)}}=[{\mathbf{h}_{v_1}^{(l)}}; {\mathbf{h}_{v_2}^{(l)}}; ...;{\mathbf{h}_{v_{L_2}}^{(l)}}]^T \in \mathbb{R}^{L_2 \times D}$ with $L_2$ being the number of nodes. The above calculation process is formulated as follows:
\begin{align}
\mathbf{H}_{\vdash}^{(l)} &= \hat{\mathbf{D}_{\vdash}}^{-1/2} \hat{\mathbf{A}_{\vdash}}\hat{\mathbf{D}_{\vdash}}^{-1/2} \mathbf{H}^{(l-1)} \mathbf{W}_{\vdash}^{(l-1)} \\
\mathbf{H}_{\dashv}^{(l)} &= \hat{\mathbf{D}_{\dashv}}^{-1/2} \hat{\mathbf{A}_{\dashv}}\hat{\mathbf{D}_{\dashv}}^{-1/2} \mathbf{H}^{(l-1)} \mathbf{W}_{\dashv}^{(l-1)} \\
\mathbf{H}^{(l)} &= \sigma(\text{concat}(\mathbf{H}_{\vdash}^{(l)}, \mathbf{H}_{\dashv}^{(l)})\mathbf{W}_f^{(l-1)})
\end{align}
where $\hat{\mathbf{A}_{\vdash}} = \mathbf{A}_{\vdash} + \mathbf{I}$ and $\hat{\mathbf{A}_{\dashv}} = \mathbf{A}_{\dashv} + \mathbf{I}$ represent the incoming and outgoing adjacency matrices of $\mathcal{G}_2$ with self-loops. And $\mathbf{I}$ is an identity matrix representing the self-loop of each node. $\hat{\mathbf{D}}_{\vdash, ii} = \sum_{j=1}^{L_2} \hat{\mathbf{A}}_{\vdash, ij}$ and $\hat{\mathbf{D}}_{\dashv, ii} = \sum_{j=1}^{L_2} \hat{\mathbf{A}}_{\dashv, ij}$ are the diagonal degree matrices.  $\mathbf{W}_{\vdash}^{(l-1)}$, $\mathbf{W}_{\dashv}^{(l-1)}$ and $\mathbf{W}_f^{(l-1)}$ denote trainable weight matrices. $\sigma$ is a non-linearity function and $\text{concat}(\cdot)$ is the concatenation operation. $L_2$ is the number of nodes in $\mathcal{G}_2$. Similar to the bi-GMP encoder, $\mathbf{H}^{(0)}$ is also initialized with GloVe embeddings. Bidirectional node embeddings at the same layer are fused by concatenation and fed into a one-layer feed-forward neural networks before being fed to the next bi-GCN layer. We stack the bi-GCN encoder to $L$ layer and $R_{2}=\mathbf{H}^{(L)}$ is the set of final node representations after $L$ hops of bi-GCN computation.

Similarly, we apply max pooling on $R_{2}$ to calculate the graph embedding of graph $\mathcal{G}_2$: 
\begin{equation}
    Z_{\mathcal{G}_2} = \text{maxpool}(R_2).
\end{equation}
where $R_2 \in \mathbb{R}^{L_2 \times D}$ and $Z_{\mathcal{G}_2} \in \mathbb{R}^D$.

\subsection{Dual Encoders Combination Strategy}
To jointly learn the global and local structural information of the RDF triples input, we propose a strategy to combine the bi-GMP encoder and the bi-GCN encoder. 

Both encoders produce a set of node representations. 
$R_{1}$ is mainly concerned with the global structural information between the triples, because the bi-GMP encoder calculates hidden states according to the traversal order obtained by the combination of topological sort algorithm and single-source shortest path algorithm. Therefore, each hidden state $\mathbf{r}_i$ can retain information from one triple to multiple triples if there exists a meta-path.
At the same time, $R_{2}$ better models the local structural information within the triples because each node representation is directly learned by all its one-hop neighbors. We stack the bi-GCN encoders to 2 layers and the output of bi-GCN encoder $\mathbf{h}_i$ retains at most two-hop information of the graph, i.e., it is limited to one or two triples.

At last, the combined graph embedding $Z_\mathcal{G}$ is computed as follows:
\begin{equation}
    Z_\mathcal{G}=Z_{\mathcal{G}_1} \oplus Z_{\mathcal{G}_2}
\end{equation}
where $\oplus$ denotes component-wise addition and $Z_\mathcal{G} \in \mathbb{R}^D$. 

\subsection{Decoder}
An attention-based LSTM decoder~\cite{luong2015effective} is employed for text generation. Since we have two encoder outputs, the conventional top-down attention cannot fully capture different semantic information from two completely different encoders. To better decode the text description, we first apply separated attention mechanism on both graph encoders and then fuse the context vectors {using a selection mechanism inspired by pointer-generator \cite{see2017get}}.

The decoder takes the combined graph embedding $Z_\mathcal{G}$ as its initial hidden state. At each decoding time step $t$, the embedding of the current input (or previously generated word) $\mathbf{e}_t$ and previous hidden state $\mathbf{s}_{t-1}$ are fed into the decoder to update its hidden state $\mathbf{s}_t$:
\begin{equation}
    \mathbf{s}_t = \text{LSTM}(\mathbf{e}_t, \mathbf{s}_{t-1})
\end{equation}

Next, we apply the separated attention mechanism by calculating the attention align weights at this time step as follows:
\begin{equation}
\alpha _{t(i)} = \frac{\exp(score( {\mathbf{r}_i}, \mathbf{s}_t))}{\sum_{k=1}^{L_1} \exp(score({\mathbf{r}_{k}}, \mathbf{s}_t))} \\
\end{equation}
\begin{equation}
\beta _{t(j)} = \frac{\exp(score( {\mathbf{h}_j}, \mathbf{s}_t))}{\sum_{k=1}^{L_2} \exp(score({\mathbf{h}_{k}}, \mathbf{s}_t))}
\end{equation}
where $\mathbf{r}_i \in R_{1}$ and $\mathbf{h}_j \in R_{2}$. $L_1=|R_{1}|$ and $L_2=|R_{2}|$ are the lengths of $R_{1}$ and $R_{2}$, respectively. $\mathbf{s}_t$ is the hidden state of decoder. In this paper, we use the $score(\cdot)$ function as follows to compute the similarity of $\mathbf{r}_i, \mathbf{h}_j$ and $\mathbf{s}_t$:
\begin{equation}
    score(\mathbf{x}, \mathbf{s}_t) = \mathbf{v} \tanh (\mathbf{W}_x \mathbf{x}+\mathbf{W}_s \mathbf{s}_t)
\end{equation}
where $\mathbf{v}, \mathbf{W}_x$ and $\mathbf{W}_s$ are learnable parameters. $\mathbf{x}$ can be $\mathbf{r}_i$ or $\mathbf{h}_j$. After obtaining the attention align weights, we calculate the context vector $\mathbf{c}_u$ of bi-GMP and the context vector $\mathbf{c}_v$ of bi-GCN respectively:
\begin{align}
\mathbf{c}_u &= \sum_{i=1}^{L_1} \alpha_{t(i)} \mathbf{r}_i \\
\mathbf{c}_v &= \sum_{j=1}^{L_2} \beta_{t(j)} \mathbf{h}_j
\end{align}

Selection mechanism uses a \textit{gcn probability} to control the attention paying to the context vectors of different encoders from a global perspective.
The \textit{gcn probability} $p_{gcn} \in [0,1]$ for time step $t$ is calculated from the decoder state $\mathbf{s}_t$ and the decoder input $\mathbf{e}_t$: 
\begin{equation}
    p_{gcn} = \sigma (\mathbf{W}_p[\mathbf{s}_t; \mathbf{e}_t]+\mathbf{b}_p)
\end{equation}
where $\mathbf{W}_p$ and $\mathbf{b}_p$ are learnable parameters, $\sigma(\cdot)$ is the sigmoid function and $[\cdot;\cdot]$ denotes concatenation operation. The fused vector $\mathbf{c}_t$ for time step $t$ combining context vectors $\mathbf{c}_u$ and $\mathbf{c}_v$ is computed as:
\begin{equation}
    \mathbf{c}_t = (1-p_{gcn})\mathbf{c}_u + p_{gcn} \mathbf{c}_v
\end{equation}
where $p_{gcn}$ is used to control the proportion of different context vectors. The final attentional hidden state at this time step is:
\begin{equation}
    \widetilde{\mathbf{s}_t} = \tanh(\mathbf{W}_c \cdot [\mathbf{c}_t; \mathbf{s}_t]+\mathbf{b}_c)
\end{equation}
where $\mathbf{W}_c$ and $\mathbf{b}_c$ are learnable parameters. 

After obtaining an attentional decoder hidden states sequence $\left \langle \widetilde{\mathbf{s}_1}, \widetilde{\mathbf{s}_2}, ... , \widetilde{\mathbf{s}_T} \right\rangle $ recurrently, the decoder generates an output sequence $\left \langle y_1, y_2, ... , y_T \right\rangle$ with $T$ being the length of the sequence. The output probability distribution over the vocabulary of the current time step is calculated by:
\begin{equation}
    P(y_t|y_{1:t-1})=softmax(\mathbf{W}_v\widetilde{\mathbf{s}_t}+\mathbf{b}_v)
\end{equation}
where $\mathbf{W}_v$ and $\mathbf{b}_v$ are trainable parameters. Finally, the text generation loss is a cross entropy loss:
\begin{equation}
    L_{g} =\frac{1}{T} \sum_{t=1}^T -\log{P(y_t^*|S, y_{1:t-1}^*)}\label{gloss} 
\end{equation}
where $S$ is the collection of input RDF triples and $y_t^*$ is the $t$-th word in the ground truth output sequence.

\begin{table*}[]
\centering
\begin{tabular}{|l|l|}
\hline
{Ground Truth}  & 1. (Alan Shepard , timeInSpace , 130170 minutes) \\ 
Triples & 2. (Alan Shepard , birthPlace , New Hampshire) \\ 
& 3. (New Hampshire , bird , purple finch)  \\ \hline
Ground Truth  & Alan Shepard was born in New Hampshire , the home of the purple finch , \\ 
Text & and spent 130170 minutes in space . \\
\hline
Generated Text  & Alan Shepard was born in New Hampshire , where the purple finch is the bird .\\ \hline
{SPN-predicted} & 1. (Alan Shepard , birthPlace , New Hampshire)  \\ 
Triples & 2. (New Hampshire , bird , purple finch) \\ \hline
\end{tabular}
\caption{\label{reward example} {An example of a ground truth triples-text pair and its corresponding generated text and predicted triples.}}
\end{table*}

\begin{figure}
\centering
\includegraphics[scale=0.5]{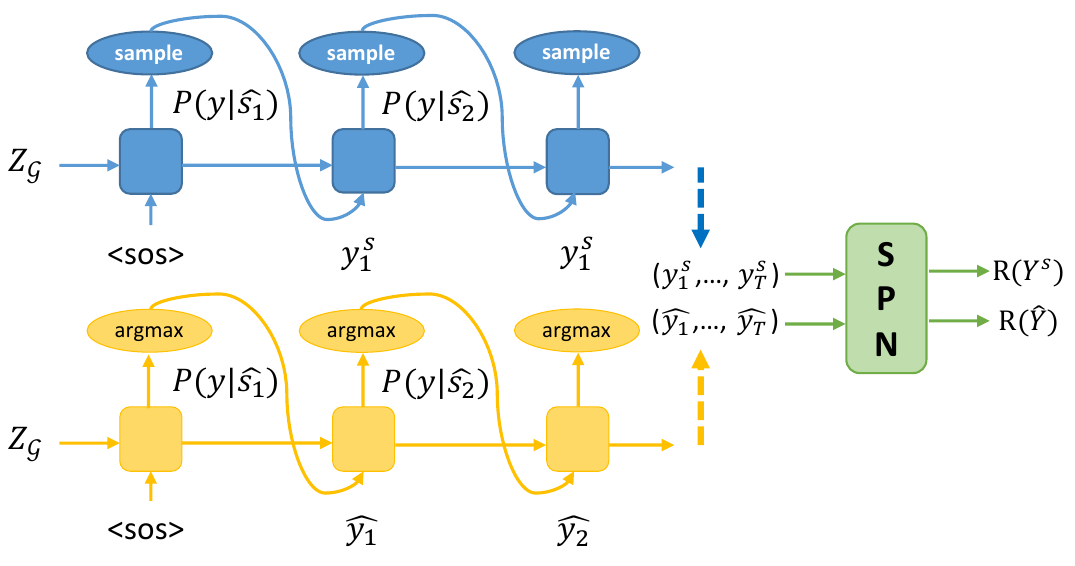}
\caption{The self-critical sequence training process of the information extraction based reinforcement learning module.
}
\label{fig:rl}
\end{figure}

\subsection{Reinforcement Learning with Information Extraction}
Besides training with the generation loss $L_g$, we further design an information extraction reward to generate more faithful and informative descriptive texts.

We apply a self-critical sequence training (SCST) algorithm \cite{rennie2017self} depicted in Fig.\ref{fig:rl} to optimize the RL loss. SCST is a form of REINFORCE algorithm that, instead of estimating a ``baseline" to normalize rewards and reduce variance, normalizes the rewards it experiences using the output of its own test-time inference algorithm. At each training iteration, two text sequences are generated: (1) a sampled sequence $Y^s$ generated by multinomial sampling, which means sampling tokens according to the probability distribution $P(y^s_t|S, y^s_{1:t-1})$; (2) a baseline sequence $\hat{Y}$ generated by greedy search. The rewards of the two text sequences $R(Y^s)$ and $R(\hat{Y})$ are defined as the correct number of predicted triples using a pretrained IE model SPN metioned in Section \ref{related work IE}. The SPN model receives the sampled sequence or the baseline sequence as input and predicts a collection of RDF triples. Tables \ref{reward example} shows a ground truth triples-text pair, a prediction text generated by our proposed model based on the ground truth triples and a collection of RDF triples predicted by the SPN model based the generated text. The IE reward for this example is equal to 2 because the generated text is missing the information in the triple (Alan Shepard , timeInSpace , 130170 minutes).
The RL objective is defined as follows:

\begin{equation}
    L_{rl} = (R(Y^s)-R(\hat{Y}))\sum_t P(y^s_t|S, y^s_{1:t-1})
\end{equation}

We train our model using a hybrid objective function combining both text generation loss and RL loss, defined as:

\begin{equation}
    L = \gamma L_{rl}+(1-\gamma)L_g
\end{equation}
where a scaling factor $\gamma$ controls the trade-off between generation loss and RL loss. 




\begin{figure}
\centering
\includegraphics[scale=0.28]{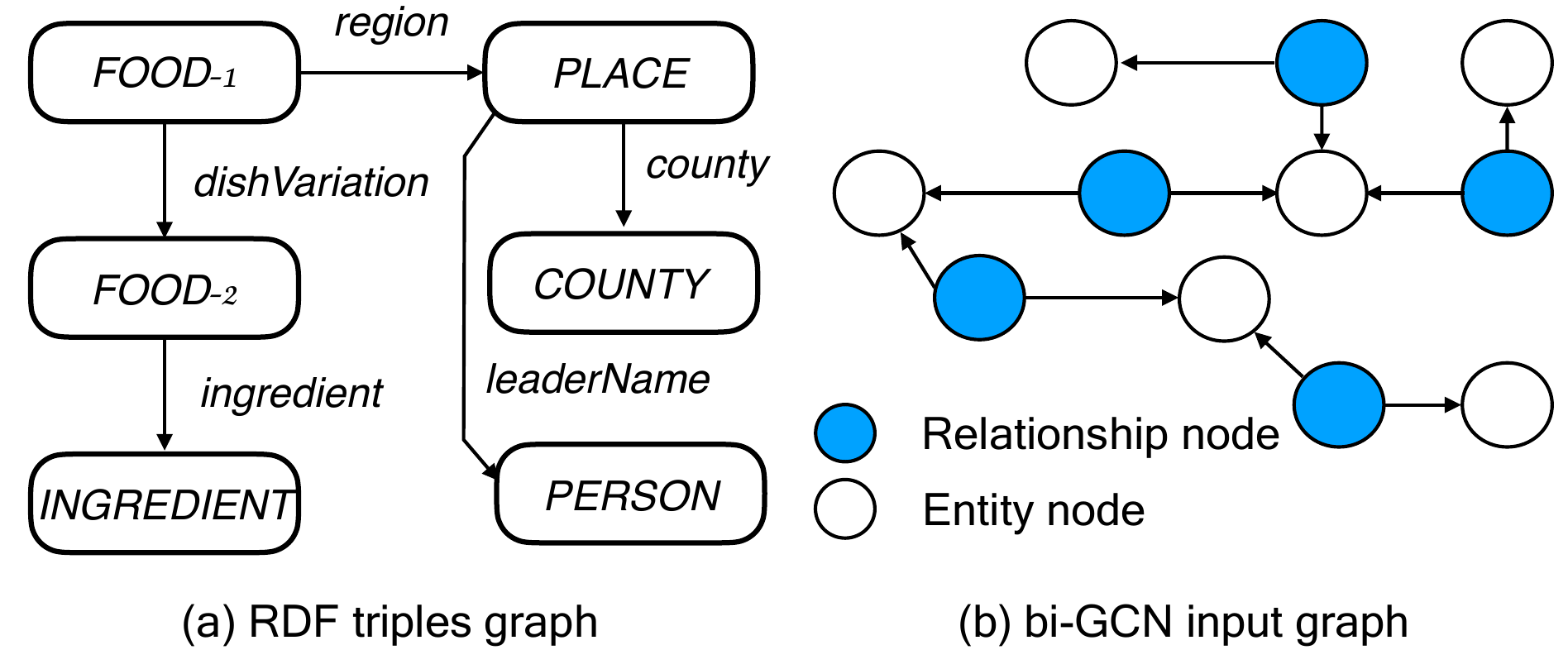}
\caption{Examples of two graph inputs. 
An RDF triples graph is shown in (a). In (b), entities (such as ``\textit{FOOD-1}'') are converted to entity nodes and relationships (such as ``\textit{region}'') are converted as new set of nodes to construct input graph for bi-GCN.
}
\label{fig:graph-constructions}
\end{figure}

\section{Entity Masking and Graph Constructions}\label{s:Graph Constructions}
In this section, we first introduce the technique of entity masking and its significance. Then, we describe how to convert a collection of RDF triples to the input graphs for two encoders.

\subsection{Entity Masking} 
\label{Entity-Masking}
Entity masking is, as the name implies, replacing the entity names that appear in the input triples and output texts with their corresponding types. It can improve the generalization ability of the models. In this paper, we use the official dictionary provided by WebNLG. Taking into account that entities appearing in the same collection of triples may of the same type, we allocate an entity-id (eid) to each entity mention in the same collection. In this way, each entity is replaced with its eid and its entity type. For instance, the entity ``\textit{Bakewell pudding}'', namely ``\textit{FOOD-1}'' in Fig. \ref{fig:graph-constructions}, is replace by ``\textit{ENTITY-1 FOOD}''. 

\subsection{RDF Triples to Meta-paths}
As shown in Fig. \ref{fig:graph-constructions}(a), to find all meta-paths, we first build the zero in-degree nodes set $\mathcal{V}_{IN}$ = \{\textit{FOOD-1}\} and the zero out-degree nodes set $\mathcal{V}_{OUT}$ = \{\textit{PERSON, COUNTY, INGREDIENT}\}. Then, we need to find the single source shortest paths from nodes in $\mathcal{V}_{IN}$ to nodes in $\mathcal{V}_{OUT}$. The three meta-paths of this example are as follows: 
\begin{itemize}
\item \textit{FOOD-1${\rightarrow}$region${\rightarrow}$ PLACE${\rightarrow}$leaderName${\rightarrow}$PERSON}
\item \textit{FOOD-1${\rightarrow}$region${\rightarrow}$PLACE${\rightarrow}$county${\rightarrow}$COUNTY}
\item \textit{FOOD-1${\rightarrow}$dishVariation${\rightarrow}$FOOD-2${\rightarrow}$ingredient${\rightarrow}$INGREDIENT}
\end{itemize}
These meta-paths are concatenated to a text sequence and input to the bi-GMP encoder. As shown in the bi-GMP module of Fig. \ref{fig:combined_modle}, the hidden state of the last token in path $p_t$ is not passed to the first token in path $p_{t+1}$. 

\subsection{RDF Triples to bi-GCN Graph}
\label{RDF Triples to bi-GCN Graph}
Similar to \cite{marcheggiani2017encoding}, the relationships are regarded as additional nodes for bi-GCN encoder and the new relationship node connects to the subject and object through two new directed edges, respectively. Therefore, the new graph  (Fig.  \ref{fig:graph-constructions}(b)) is two times larger than the original RDF triples graph (Fig.  \ref{fig:graph-constructions}(a)) in hops, increasing the difficulty to capture long-range dependency. 
The input triple (\textit{FOOD-1, region, PLACE}) is split into two new connections ``\textit{region$\rightarrow$FOOD-1}'' and ``\textit{region$\rightarrow$PLACE}'' with the relationship ``\textit{region}'' being a new node in the input graph. For nodes with multiple tokens, each token is split into a separate node with a new edge connecting to the root node (usually the ``\textit{ENTITY-id}'' node).
For example, the original entity ``\textit{ENTITY-1 FOOD INGREDIENTS}'' is split into two connections ``\textit{ENTITY-1$\rightarrow$FOOD}'' and ``\textit{ENTITY-1$\rightarrow$INGREDIENTS}''.

\section{Experiments}
\subsection{Datasets}
We use two open RDF-to-text generation datasets WebNLG \footnote{\url{https://gitlab.com/shimorina}} \cite{gardent2017creating} and DART \footnote{\url{https://github.com/Yale-LILY/dart}} \cite{nan2021dart} to evaluate our proposed model.
Each example in the dataset is a (triples, text) pair and one triple collection can correspond to multiple ground truth texts. An RDF triple is composed of (subject, relationship, object), in which the subject and object are either constants or entities. The WebNLG 2017 challenge dataset includes 18102 training pairs, 2268 validation pairs, and 2495 test pairs from 10 categories (Airport, Astronaut, City, Building, ComicsCharacter, Food, Monument, SportsTeam, University, WrittenWork). {In addition to being divided by entity categories, the WebNLG dataset is also divided into ``seen", ``unseen" and ``all" categories based on whether the topic entities have been exposed in the training set.}

{The second dataset DART is a large and open-domain structured \textbf{DA}ta \textbf{R}ecord to \textbf{T}ext generation corpus collection, integrating data from WikiSQL, WikiTableQuestions, WebNLG 2017 and Cleaned E2E. The DART dataset is composed of 62659  training  pairs,  6980  validation  pairs,  and  12552  test  pairs and covers more categories since the WikiSQL and WikiTableQuestions datasets come from open-domain Wikipedia. Unlike the previous dataset, this DART dataset is not classified as ``seen'' and ``unseen'' and therefore will not be performed entity masking.}

\subsection{Experimental Settings and Evaluation Metrics} 
The vocabulary is built based on the training set, shared by the encoders and decoder. 
{We implement our proposed model based on Graph4NLP\footnote{\url{https://github.com/graph4ai/graph4nlp}} released by ~\cite{wu2021graph}. }
For the hyperparameters of the model, we set 300-dimensional word embeddings and a 512-dimensional hidden state for the bi-GCN encoder, bi-GMP encoder and decoder. Adam~\cite{kingma2014adam} is used as the optimization method with an initial learning rate 0.001 and the batch size is 50. 
The scaling factor $\gamma$ controlling the trade-off between generation loss and RL loss is 0.3.

As for the evaluation metrics, we use the standard evaluation metrics of the WebNLG challenge, including BLEU~\cite{papineni2002bleu}, METEOR~\cite{denkowski2011meteor} and TER. 
BLEU looks at n-gram overlaps between the output and ground truth text with a penalty for shorter outputs. {The METEOR aligns hypotheses to one or more reference translations based on exact, stem, synonym, and paraphrase matches between words and phrases. Translation Error Rate (TER) is an automatic character-level measure of the number of editing operations required to translate machine translated output into human translated reference.}
The metric of BLEU provided by the WebNLG 2017 challenge is multi-BLEU\footnote{The script of multi-bleu is available on \url{https://github.com/moses-smt/mosesdecoder/blob/master/scripts/generic/multi-bleu.perl}}, which is a weighted average of BLEU-1, BLEU-2, BLEU-3 and BLEU-4.
For BLEU and METEOR, the higher the better. For TER, the lower the better. 

\subsection{Baselines}
Our proposed combined models is compared against the following baselines:
\paragraph{Sequential Model} \textsc{PKUWriter}, which is reported in \cite{gardent2017webnlg}, is a sequential model containing an attention-based encoder-decoder framework based on the TensorFlow seq2seq model.
The RDF triples are converted into a sequence and fed into the encoder. 

\paragraph{GTR-LSTM} \citeauthor{trisedya2018gtr} \cite{trisedya2018gtr} proposed a novel graph-based triple encoder to retain the information from the collection of RDF triples.

\paragraph{DGCN} \citeauthor{marcheggiani2018deep} \cite{marcheggiani2018deep} proposed to use graph convolutional networks to directly exploit the input structure. 

\paragraph{Transformer} \citeauthor{ferreira2019neural} \cite{ferreira2019neural} introduced a comparison between neural pipeline and end-to-end approaches for RDF2Text generation using Gated-Recurrent Units
(GRU) and Transformer.

\paragraph{Bi-GMP Model} Bi-GMP model contains a meta-paths encoder computing the token representations in the meta-paths from both directions and a one-layer attention-based LSTM decoder.

\paragraph{Bi-GCN Model} Bi-GCN model contains a two-layer bi-GCN encoder and a one-layer attention-based LSTM decoder. 

\begin{table*}[]
\centering
\small
\begin{tabular}{cccccccccc}
\toprule[1pt]
\multirow{2}{*}{Model} & BLEU & BLEU & BLEU & METEOR & METEOR & METEOR & TER & TER & TER \\
 & seen $\uparrow$ & unseen $\uparrow$ & all $\uparrow$  & seen $\uparrow$ & unseen $\uparrow$ & all $\uparrow$ & seen $\downarrow$ & unseen $\downarrow$ & all $\downarrow$ \\ \hline
\textsc{PKUWriter}  & 51.23 & 25.36 & 39.88 & 0.37 & 0.24 & 0.31 & 0.45 & 0.67 & 0.55 \\
\textsc{GTR-LSTM}  & 54.00 & \textbf{29.20} & 37.10 & 0.37 & \textbf{0.28} & 0.31 & 0.45 & \textbf{0.60} & 0.55 \\
\textsc{DGCN}  & 55.90 & - & - & 0.39 & - & - & 0.41 & - & - \\
Transformer & 56.28 & 23.04 & 42.41 & \textbf{0.42} & 0.21 & 0.32 & \textbf{0.39} & 0.63 & \textbf{0.50} \\
bi-GCN & 57.01 & 26.82 & 43.62 & 0.40 & 0.25 & 0.33 & 0.42 & 0.68 & 0.54 \\ 
bi-GMP & 57.25 & 24.03 & 43.32 & 0.40 & 0.22 & 0.32 & 0.43 & 0.63 & 0.52 \\ 
\hline 
Ours & \textbf{57.96} & 24.70 & 43.62 & 0.41 & 0.23 & 0.33 & 0.40 & 0.65 & 0.52 \\
Ours + IE & 57.79 & 26.55 & \textbf{44.00} & 0.41 & 0.26 & \textbf{0.34} & 0.41 & 0.66 & 0.53 \\
\hline
\toprule[1pt]
\end{tabular}
\caption{\label{result-table} {BLEU, METEOR and TER on WebNLG 2017 challenge test dataset. Bold fonts indicate the best results. \textbf{Ours} indicates the bi-GCN + bi-GMP model.}}
\end{table*}

\begin{table}[]
\centering
\begin{tabular}{cccc}
\toprule[1pt]
Model & BLEU & METEOR & TER \\ \hline
End-to-End Transformer & 27.24 & 0.25 & 0.65\\
Seq2Seq-Att & 29.66 & 0.27 & 0.63 \\
bi-GCN & 38.73 & \textbf{0.32} & 0.6 \\ 
bi-GMP & 39.43 & \textbf{0.32} & \textbf{0.57} \\ 
\hline 
bi-GCN + bi-GMP & 39.7 & \textbf{0.32} & \textbf{0.57} \\
bi-GCN + bi-GMP + IE & \textbf{39.87} & \textbf{0.32} & \textbf{0.57} \\ \hline
\toprule[1pt]
\end{tabular}
\caption{\label{result-table-dart} BLEU, METEOR and TER on DART test dataset. The results of End-to-End Transformer and Seq2Seq-Att are reported in \cite{nan2021dart}. }
\end{table} 

\begin{table*}[]
\centering
\begin{tabular}{c|cc|cc}
\toprule[1pt]
\multirow{2}{*}{\diagbox{Model}{Dataset}}     & \multicolumn{2}{c|}{WebNLG 2017 Challenge} & \multicolumn{2}{c}{DART} \\ \cline{2-5} &
   $1 \leq \text{size} \leq 3$  &   
   $4 \leq \text{size} \leq 7$  &   
   $1 \leq \text{size} \leq 3$  &   
   $4 \leq \text{size} \leq 7$   
   \\ \hline
bi-GCN & 46.99 & 40.37 & 32.00 & 36.41\\ 
bi-GMP & 46.81 & 40.32 & 33.56 & 37.13\\
bi-GCN + bi-GMP & 47.10 & \textbf{40.62} & 34.50 & 37.07\\
bi-GCN + bi-GMP + IE & \textbf{48.20} & 40.45 & \textbf{34.75} & \textbf{37.15}\\
\toprule[1pt]
\end{tabular}
\caption{\label{bleu-table-size} {The BLEU scores of the baseline models and our proposed models according to different triple set sizes.} }
\end{table*}

\subsection{Experimental Results} 
Table \ref{result-table} shows the experimental results on the WebNLG test dataset. Our proposed combined models achieve best results on BLEU-seen, BLEU-all and METEOR-all metrics, and also achieve comparable results on other evaluation metrics. The series of models implemented under our framework including bi-GCN, bi-GMP, bi-GCN+bi-GMP, and bi-GCN+bi-GMP+IE are 0.7 to 1.7 points and 0.9 to 1.6 points higher in BLEU-seen and BLEU-all respectively than the previous baseline models. The bi-GCN model has lower score than bi-GMP on BLEU-seen, but higher score on BLEU-unseen than bi-GMP, as well as higher score than some sequence models such as Transformer and \textsc{PKUWriter}, showing the better generalization ability of graph neural network model for the unseen categories.

Table \ref{result-table-dart} shows the experimental results on the DART test dataset. As shown in this table, our proposed combined models consistently outperform other baselines on all three evaluation metrics. This is because our full model could better capture the global and local graph structure of the RDF triples. The performances of the models implemented under our framework significantly surpass the sequential baselines by about 10 BLEU points. The TER score (0.6) of bi-GCN model is higher than those (0.57) of other models proposed by us, which also shows a similar trend in the WebNLG dataset. We believe that this may be because the text generated by the graph neural network states the information in the triple set in a more random order. 

Further experiments are carried out to verify the effectiveness of reinforcement learning with information extraction in Table \ref{bleu-table-size}. Since each example in the dataset may contain from one to multiple triples (up to seven triples for WebNLG and more for DART), we divide each dataset into two subsets with less than or equal to three and more than three triples according to the number of triples. In particular, for the DART dataset in this table, we only consider examples containing up to seven triples. As shown in this table, BLEU-all score gains 1.1 points on $1 \leq \text{size} \leq 3$ subset of the WebNLG dataset after introducing the IE-based reinforcement learning reward. For the DART dataset, the introduction of IE-based RL reward results in a further improvement in BLEU-all scores for both subsets, and the improvement is more significant in the $1 \leq \text{size} \leq 3$ subset. This is because the information extraction model performs better in examples where the number of input triples is small. When the number of triples contained in the input text is large, it is difficult for the IE module to extract them completely, and the error propagation from extraction may also lead to limited improvement of the model.

\begin{table}[]
\begin{tabular}{c|ccccc}
\toprule[1pt]
num of triples     & 1      & 2      & 3      & 4      & 5      \\ \hline
ground truth         & 0.656 & 0.612 & 0.529 & 0.482 & 0.431 \\
bi-GCN + bi-GMP      & 0.517 & 0.482 & 0.409 & 0.359 & 0.326 \\
bi-GCN + bi-GMP + IE & 0.558 & 0.503 & 0.422 & 0.355 & 0.323 \\
\toprule[1pt]
\end{tabular}
\caption{Averaged F1 score of questions answered correctly for each generated text of WebNLG with FEQA, a QA-based metric for summary faithfulness proposed by \cite{durmus2020feqa}.}
\label{feqa_webnlg_f1}
\end{table}

Moreover, we introduce the automatic evaluation metric FEQA \cite{durmus2020feqa} used in the field of abstractive summarization to evaluate the faithfulness of generated summmaries. Given question answer pairs generated from the summary, FEQA extracts answers from the document; non-matched answers indicate unfaithful information in the summary. In our setting, we regard the sequence of input triples as a document and regard the generated text as a summary. FEQA uses the  averaged F1 score of questions answered correctly for each generated text. Table \ref{feqa_webnlg_f1} shows the experimental results on WebNLG test dataset. We split the dataset by the size of input triples and perform FEQA evaluation on each split. The ground truth texts, texts generated by bi-GCN+bi-GMP model, and texts generated by bi-GCN+bi-GMP+IE model are evaluated. As the table shown, the F1 scores of ground truth are the highest and the F1 scores of bi-GCN+bi-GMP+IE model are higher than bi-GCN+bi-GMP model when the size is smaller than 4, which is consistent with the results in Table \ref{bleu-table-size} and verifies the effectiveness of our proposed IE-based reinforment learning reward. In particular, we do not perform FEQA on splits where the input size is larger than 5 triples, as these generated or ground truth texts are so informative that it is difficult for the model to generate reasonable questions.
Our code is publicly available for research purpose. 
\footnote{\url{https://github.com/Nicoleqwerty/RDF-to-Text}.}


\begin{table*}[ht]
\begin{tabu}to \hsize {XX[8,l]}
\toprule[1pt]
Model & Text \\ \hline
\Blue{RDF Triples} &\Blue{(turkey, leaderName, ahmet davutoglu), (turkey, capital, ankara), (turkey, largestCity, istanbul), (ataturk monument ( izmir ), material, bronze), (turkey, currency, turkish lira), (ataturk monument ( izmir ), inaugurationDate, 1932 - 07 - 27), (ataturk monument ( izmir ), location, turkey)} \\ \hline
{Reference}  & {the lira is the official currency of turkey where ahmet davutoglu is the leader . although the largest city is istanbul the capital city is ankara . the country is the location of the bronze ataturk in izmir which was inaugurated on 27 july 1932 .}\\ \hline
\Black{bi-GCN} & \Black{the ataturk monument is located in izmir , turkey , where the capital is ankara and the leader is ahmet davutoglu . the currency of turkey is the lira and \textit{the leader is ahmet davutoglu .}} \\ \hline
\Black{bi-GMP} & \Black{the ataturk monument in izmir was inaugurated on 27 july 1932 in \textbf{izmir} . the capital of turkey is ankara and the leader is ahmet davutoglu .}\\ \hline

\Green{bi-GCN + bi-GMP} & \Green{the ataturk monument in izmir , turkey , where the capital is ankara , \textit{is ahmet davutoglu} . the leader of turkey is ahmet davutoglu and the currency is the lira .}\\ \hline
\Green{bi-GCN + bi-GMP + IE} & \Green{the ataturk monument in izmir , turkey , which is made of bronze , was inaugurated on 27 july 1932 . the capital of turkey is ankara and the currency is the lira .}\\ \hline
\end{tabu}
\caption{\label{output-table} An example of triples input, reference and outputs of different models. The bold tokens are incorrect outputs. The italic tokens indicate repeated outputs.}
\end{table*}

\begin{table*}[ht]
\begin{tabu}to \hsize {X|X[8,l]|X}
\toprule[1pt]
& Text  & BLEU-4 \\ \hline
triples & (adams county , pennsylvania, has to its southeast, carroll county , maryland)  &  \\
ours  & carroll county , maryland is \textbf{to the southwest} of adams county , pennsylvania . & 0.5337 \\
ours+ie & adams county , pennsylvania has carroll county , maryland to its southeast .  & 0.9999 \\
reference  & to the southeast of adams county , pennsylvania lies carroll county , maryland . & \\ \hline
triples   & (antioch , california, elevationAboveTheSeaLevel, 13 . 0), (antioch , california, areaTotal, 75 . 324) &  \\ 
ours    & antioch , california is 13 . 0 square kilometres and \textbf{is located at 13 . 0 }.& 0.3086 \\ 
ours+ie & antioch , california is 13 . 0 metres above sea level and has a total area of 75 . 324 square kilometres .  & 0.7801 \\ 
reference  & antioch , california is 13 . 0 metres above sea level and has a total area of 75 . 324 sq km .   & \\ 
\toprule[1pt]
\end{tabu}
\caption{\label{case-study-size} Two examples with one input triple and two input triples respectively, and their corresponding GCN+GMP, GCN+GMP+IE model outputs and ground truth reference outputs. The bold tokens are incorrect outputs.}
\end{table*}

\subsection{Ablation Study} 
Table \ref{result-table} and Table \ref{result-table-dart} indicate that there are three key factors that may affect the generated text quality. 
The first two factors are the bi-GCN and bi-GMP encoders.
Experimental results show that the performance of combined graph-augmented structural neural encoders is better than that of the models with a single graph-based encoder.
This result is as expected because it is tough for one single graph-based encoder to completely model both global and local structural information.
The third factor is information extraction based reinforcement learning reward. The introduction of this reward helps the model generate more faithful texts and the improvement is particularly noticeable in the examples with less than 4 triples.

\subsection{Case Study}
Furthermore, we manually examine the output of different models and perform case studies to better understand the performance of different models. 
The triples for example in Table \ref{output-table} is described around the entity ``turkey'' (involving 4 triples) and the entity ``ataturk monument'' (involving 3 triples).
As shown in this table, we find that the models that include the bi-CCN encoder perform better in correctly predicting the relationships between entities.
This is expected because the bi-GCN encoder focuses more on the local structural information and can predict relationships within triples more efficiently and accurately. However, the bi-GCN encoder cannot capture long-range dependency well, which mainly focuses on the local information around the entity ``turkey'' and may introduce some repeated information.
Meanwhile, the bi-GMP encoder mainly concentrates on the relationships between triples and can help the full model to cover different aspects of information. In the example in Table \ref{output-table}, the bi-GMP encoder pays almost equal attention to the entity ``turkey'' and entity ``ataturk monument''. But the bi-GMP model tends to generate incorrect triple information (monument, location, izmir) and short text. 
In general, our proposed models can cover more information/triples and generate more accurate texts.

{As the two examples in the Table \ref{case-study-size} show, the bi-GCN+bi-GMP+IE model does help to improve the faithfulness of the generated text, especially for examples with less than 4 triples. The model without IE-based RL reward may incorrectly predict relationships, such as incorrectly predicting ``has to its southeast'' as ``to the southwest''. It may also miss some important triple information, such as missing the triple (antioch california, areaTotal, 75.324).}

\begin{table}[]
\centering
\begin{tabular}{cccc}
\toprule[1pt]
\multirowcell{1}{Methods} & \multirowcell{1}{Grammar} & \multirowcell{1}{Informativity} & \multirowcell{1}{Faithfulness} \\ \hline
Reference  & 4.46 & 4.82 & 4.65  \\ 
bi-GMP  & 4.26 & 4.09 & 4.27 \\ 
bi-GCN & 4.21 & 3.98 & 4.44 \\ 
Our Full Model & 4.55 & 4.24 & 4.57  \\ 
\toprule[1pt]
\end{tabular}
\caption{\label{human-eval-table} Human evaluation of WebNLG 2017 Challenge test dataset.
The higher the score is, the better the performance is. }
\end{table}

\subsection{Human Evaluation}
In order to further assess the quality of these generated texts, we handed some original RDF triples and their generated text pairs to three human evaluators. The three evaluators were gievn 200 outputs, of which 50 were reference texts, 50 were outputs of the bi-GCN model, 50 were outputs of the bi-GMP model, and 50 were outputs of the full model. They were requested to assess the generated texts from three perspectives, with each perspective rated from 1 to 5. The first perspective is \textit{Grammatical}, which scores each sample in terms of coherence, absence of redundancy, and absence of grammatical errors. The second perspective is \textit{Informativity (Global)}, which evaluates how well the information appearing in input triples are covered. The third perspective is \textit{Faithfulness (Local)}, which assesses the accuracy of the entities and corresponding relationships that appear in the generated texts. We averaged the results given by the three evaluators. Table \ref{human-eval-table} indicates that our proposed model can generate more informative and faithful texts. 

\section{Conclusion}
In this paper, we propose a novel framework which explores a {reinforcement learning based} graph-augmented structural neural encoders for RDF-to-text generation. Our approach learns local and global structural information jointly through a combination of bidirectional GCN encoder and bidirectional GMP encoder.
{We further exploit a reinforcement learning reward computed by a pretrained information extraction model for the generated text to improve the text faithfulness.}
Our future work is to further extend our proposed approach and develop a knowledge graph question answering system which can answer the question in a generated sentence.

\begin{acks}
The work is partially supported by the National Nature Science Foundation of China (No. 61976160), Technology research plan project of Ministry of Public and Security (Grant No. 2020JSYJD01), Thirteen Five-year Research Planning Project of National Language Committee (No. YB135-149) and the Self-determined Research Funds of CCNU from the Colleges' Basic Research and Operation of MOE (No. CCNU20ZT012, CCNU20TD001).
\end{acks}

\bibliographystyle{ACM-Reference-Format}
\bibliography{main}


\begin{thebibliography}{70}


\ifx \showCODEN    \undefined \def \showCODEN     #1{\unskip}     \fi
\ifx \showDOI      \undefined \def \showDOI       #1{#1}\fi
\ifx \showISBNx    \undefined \def \showISBNx     #1{\unskip}     \fi
\ifx \showISBNxiii \undefined \def \showISBNxiii  #1{\unskip}     \fi
\ifx \showISSN     \undefined \def \showISSN      #1{\unskip}     \fi
\ifx \showLCCN     \undefined \def \showLCCN      #1{\unskip}     \fi
\ifx \shownote     \undefined \def \shownote      #1{#1}          \fi
\ifx \showarticletitle \undefined \def \showarticletitle #1{#1}   \fi
\ifx \showURL      \undefined \def \showURL       {\relax}        \fi
\providecommand\bibfield[2]{#2}
\providecommand\bibinfo[2]{#2}
\providecommand\natexlab[1]{#1}
\providecommand\showeprint[2][]{arXiv:#2}

\bibitem[\protect\citeauthoryear{Androutsopoulos, Lampouras, and
  Galanis}{Androutsopoulos et~al\mbox{.}}{2013}]%
        {androutsopoulos2013generating}
\bibfield{author}{\bibinfo{person}{Ion Androutsopoulos},
  \bibinfo{person}{Gerasimos Lampouras}, {and} \bibinfo{person}{Dimitrios
  Galanis}.} \bibinfo{year}{2013}\natexlab{}.
\newblock \showarticletitle{Generating natural language descriptions from OWL
  ontologies: the NaturalOWL system}.
\newblock \bibinfo{journal}{\emph{Journal of Artificial Intelligence Research}}
   \bibinfo{volume}{48} (\bibinfo{year}{2013}), \bibinfo{pages}{671--715}.
\newblock


\bibitem[\protect\citeauthoryear{Bahdanau, Cho, and Bengio}{Bahdanau
  et~al\mbox{.}}{2014}]%
        {bahdanau2014neural}
\bibfield{author}{\bibinfo{person}{Dzmitry Bahdanau},
  \bibinfo{person}{Kyunghyun Cho}, {and} \bibinfo{person}{Yoshua Bengio}.}
  \bibinfo{year}{2014}\natexlab{}.
\newblock \showarticletitle{Neural machine translation by jointly learning to
  align and translate}.
\newblock \bibinfo{journal}{\emph{arXiv preprint arXiv:1409.0473}}
  (\bibinfo{year}{2014}).
\newblock


\bibitem[\protect\citeauthoryear{Banarescu, Bonial, Cai, Georgescu, Griffitt,
  Hermjakob, Knight, Koehn, Palmer, and Schneider}{Banarescu
  et~al\mbox{.}}{2013}]%
        {banarescu-etal-2013-abstract}
\bibfield{author}{\bibinfo{person}{Laura Banarescu}, \bibinfo{person}{Claire
  Bonial}, \bibinfo{person}{Shu Cai}, \bibinfo{person}{Madalina Georgescu},
  \bibinfo{person}{Kira Griffitt}, \bibinfo{person}{Ulf Hermjakob},
  \bibinfo{person}{Kevin Knight}, \bibinfo{person}{Philipp Koehn},
  \bibinfo{person}{Martha Palmer}, {and} \bibinfo{person}{Nathan Schneider}.}
  \bibinfo{year}{2013}\natexlab{}.
\newblock \showarticletitle{{A}bstract {M}eaning {R}epresentation for
  Sembanking}. In \bibinfo{booktitle}{\emph{Proceedings of the 7th Linguistic
  Annotation Workshop and Interoperability with Discourse}}.
  \bibinfo{publisher}{Association for Computational Linguistics},
  \bibinfo{address}{Sofia, Bulgaria}, \bibinfo{pages}{178--186}.
\newblock
\urldef\tempurl%
\url{https://aclanthology.org/W13-2322}
\showURL{%
\tempurl}


\bibitem[\protect\citeauthoryear{Banik, Gardent, Scott, Dinesh, and
  Liang}{Banik et~al\mbox{.}}{2012}]%
        {banik2012kbgen}
\bibfield{author}{\bibinfo{person}{Eva Banik}, \bibinfo{person}{Claire
  Gardent}, \bibinfo{person}{Donia Scott}, \bibinfo{person}{Nikhil Dinesh},
  {and} \bibinfo{person}{Fennie Liang}.} \bibinfo{year}{2012}\natexlab{}.
\newblock \showarticletitle{KBGen: text generation from knowledge bases as a
  new shared task}. In \bibinfo{booktitle}{\emph{Proceedings of the Seventh
  International Natural Language Generation Conference}}.
  \bibinfo{pages}{141--145}.
\newblock


\bibitem[\protect\citeauthoryear{Barzilay and Lapata}{Barzilay and
  Lapata}{2005}]%
        {barzilay2005collective}
\bibfield{author}{\bibinfo{person}{Regina Barzilay} {and}
  \bibinfo{person}{Mirella Lapata}.} \bibinfo{year}{2005}\natexlab{}.
\newblock \showarticletitle{Collective content selection for concept-to-text
  generation}. In \bibinfo{booktitle}{\emph{Proceedings of the conference on
  Human Language Technology and Empirical Methods in Natural Language
  Processing}}. \bibinfo{pages}{331--338}.
\newblock


\bibitem[\protect\citeauthoryear{Bastings, Titov, Aziz, Marcheggiani, and
  Sima’an}{Bastings et~al\mbox{.}}{2017}]%
        {bastings2017graph}
\bibfield{author}{\bibinfo{person}{Jasmijn Bastings}, \bibinfo{person}{Ivan
  Titov}, \bibinfo{person}{Wilker Aziz}, \bibinfo{person}{Diego Marcheggiani},
  {and} \bibinfo{person}{Khalil Sima’an}.} \bibinfo{year}{2017}\natexlab{}.
\newblock \showarticletitle{Graph Convolutional Encoders for Syntax-aware
  Neural Machine Translation}. In \bibinfo{booktitle}{\emph{Proceedings of the
  2017 Conference on Empirical Methods in Natural Language Processing}}.
  \bibinfo{pages}{1957--1967}.
\newblock


\bibitem[\protect\citeauthoryear{Beck, Haffari, and Cohn}{Beck
  et~al\mbox{.}}{2018}]%
        {beck2018graph}
\bibfield{author}{\bibinfo{person}{Daniel Beck}, \bibinfo{person}{Gholamreza
  Haffari}, {and} \bibinfo{person}{Trevor Cohn}.}
  \bibinfo{year}{2018}\natexlab{}.
\newblock \showarticletitle{Graph-to-Sequence Learning using Gated Graph Neural
  Networks}. In \bibinfo{booktitle}{\emph{Proceedings of the 56th Annual
  Meeting of the Association for Computational Linguistics (Volume 1: Long
  Papers)}}. \bibinfo{pages}{273--283}.
\newblock


\bibitem[\protect\citeauthoryear{Belz}{Belz}{2008}]%
        {belz2008automatic}
\bibfield{author}{\bibinfo{person}{Anja Belz}.}
  \bibinfo{year}{2008}\natexlab{}.
\newblock \showarticletitle{Automatic generation of weather forecast texts
  using comprehensive probabilistic generation-space models}.
\newblock \bibinfo{journal}{\emph{Natural Language Engineering}}
  \bibinfo{volume}{14}, \bibinfo{number}{4} (\bibinfo{year}{2008}),
  \bibinfo{pages}{431--455}.
\newblock


\bibitem[\protect\citeauthoryear{Bontcheva and Wilks}{Bontcheva and
  Wilks}{2004}]%
        {bontcheva2004automatic}
\bibfield{author}{\bibinfo{person}{Kalina Bontcheva} {and}
  \bibinfo{person}{Yorick Wilks}.} \bibinfo{year}{2004}\natexlab{}.
\newblock \showarticletitle{Automatic report generation from ontologies: the
  MIAKT approach}. In \bibinfo{booktitle}{\emph{International conference on
  application of natural language to information systems}}. Springer,
  \bibinfo{pages}{324--335}.
\newblock


\bibitem[\protect\citeauthoryear{Bruna, Zaremba, Szlam, and LeCun}{Bruna
  et~al\mbox{.}}{2013}]%
        {bruna2013spectral}
\bibfield{author}{\bibinfo{person}{Joan Bruna}, \bibinfo{person}{Wojciech
  Zaremba}, \bibinfo{person}{Arthur Szlam}, {and} \bibinfo{person}{Yann
  LeCun}.} \bibinfo{year}{2013}\natexlab{}.
\newblock \showarticletitle{Spectral networks and locally connected networks on
  graphs}.
\newblock \bibinfo{journal}{\emph{arXiv preprint arXiv:1312.6203}}
  (\bibinfo{year}{2013}).
\newblock


\bibitem[\protect\citeauthoryear{Chan and Roth}{Chan and Roth}{2011}]%
        {chan2011exploiting}
\bibfield{author}{\bibinfo{person}{Yee~Seng Chan} {and} \bibinfo{person}{Dan
  Roth}.} \bibinfo{year}{2011}\natexlab{}.
\newblock \showarticletitle{Exploiting syntactico-semantic structures for
  relation extraction}. In \bibinfo{booktitle}{\emph{Proceedings of the 49th
  Annual Meeting of the Association for Computational Linguistics: Human
  Language Technologies}}. \bibinfo{pages}{551--560}.
\newblock


\bibitem[\protect\citeauthoryear{Chen, Wu, and Zaki}{Chen
  et~al\mbox{.}}{2020}]%
        {chen2019reinforcement}
\bibfield{author}{\bibinfo{person}{Yu Chen}, \bibinfo{person}{Lingfei Wu},
  {and} \bibinfo{person}{Mohammed~J Zaki}.} \bibinfo{year}{2020}\natexlab{}.
\newblock \showarticletitle{Reinforcement learning based graph-to-sequence
  model for natural question generation}. In \bibinfo{booktitle}{\emph{ICLR}}.
\newblock


\bibitem[\protect\citeauthoryear{Cho, van Merri{\"e}nboer, Bahdanau, and
  Bengio}{Cho et~al\mbox{.}}{2014}]%
        {cho-etal-2014-properties}
\bibfield{author}{\bibinfo{person}{Kyunghyun Cho}, \bibinfo{person}{Bart van
  Merri{\"e}nboer}, \bibinfo{person}{Dzmitry Bahdanau}, {and}
  \bibinfo{person}{Yoshua Bengio}.} \bibinfo{year}{2014}\natexlab{}.
\newblock \showarticletitle{On the Properties of Neural Machine Translation:
  Encoder{--}Decoder Approaches}. In \bibinfo{booktitle}{\emph{Proceedings of
  {SSST}-8, Eighth Workshop on Syntax, Semantics and Structure in Statistical
  Translation}}. \bibinfo{publisher}{Association for Computational
  Linguistics}, \bibinfo{address}{Doha, Qatar}, \bibinfo{pages}{103--111}.
\newblock
\urldef\tempurl%
\url{https://doi.org/10.3115/v1/W14-4012}
\showDOI{\tempurl}


\bibitem[\protect\citeauthoryear{Dai, Xiao, Lyu, Dou, She, and Wang}{Dai
  et~al\mbox{.}}{2019}]%
        {dai2019joint}
\bibfield{author}{\bibinfo{person}{Dai Dai}, \bibinfo{person}{Xinyan Xiao},
  \bibinfo{person}{Yajuan Lyu}, \bibinfo{person}{Shan Dou},
  \bibinfo{person}{Qiaoqiao She}, {and} \bibinfo{person}{Haifeng Wang}.}
  \bibinfo{year}{2019}\natexlab{}.
\newblock \showarticletitle{Joint extraction of entities and overlapping
  relations using position-attentive sequence labeling}. In
  \bibinfo{booktitle}{\emph{Proceedings of the AAAI conference on artificial
  intelligence}}, Vol.~\bibinfo{volume}{33}. \bibinfo{pages}{6300--6308}.
\newblock


\bibitem[\protect\citeauthoryear{Deemter, Theune, and Krahmer}{Deemter
  et~al\mbox{.}}{2005}]%
        {deemter2005real}
\bibfield{author}{\bibinfo{person}{Kees~Van Deemter},
  \bibinfo{person}{Mari{\"e}t Theune}, {and} \bibinfo{person}{Emiel Krahmer}.}
  \bibinfo{year}{2005}\natexlab{}.
\newblock \showarticletitle{Real versus template-based natural language
  generation: A false opposition?}
\newblock \bibinfo{journal}{\emph{Computational Linguistics}}
  \bibinfo{volume}{31}, \bibinfo{number}{1} (\bibinfo{year}{2005}),
  \bibinfo{pages}{15--24}.
\newblock


\bibitem[\protect\citeauthoryear{Defferrard, Bresson, and
  Vandergheynst}{Defferrard et~al\mbox{.}}{2016}]%
        {defferrard2016convolutional}
\bibfield{author}{\bibinfo{person}{Micha{\"e}l Defferrard},
  \bibinfo{person}{Xavier Bresson}, {and} \bibinfo{person}{Pierre
  Vandergheynst}.} \bibinfo{year}{2016}\natexlab{}.
\newblock \showarticletitle{Convolutional neural networks on graphs with fast
  localized spectral filtering}. In \bibinfo{booktitle}{\emph{Advances in
  neural information processing systems}}. \bibinfo{pages}{3844--3852}.
\newblock


\bibitem[\protect\citeauthoryear{Denkowski and Lavie}{Denkowski and
  Lavie}{2011}]%
        {denkowski2011meteor}
\bibfield{author}{\bibinfo{person}{Michael Denkowski} {and}
  \bibinfo{person}{Alon Lavie}.} \bibinfo{year}{2011}\natexlab{}.
\newblock \showarticletitle{Meteor 1.3: Automatic metric for reliable
  optimization and evaluation of machine translation systems}. In
  \bibinfo{booktitle}{\emph{Proceedings of the sixth workshop on statistical
  machine translation}}. \bibinfo{pages}{85--91}.
\newblock


\bibitem[\protect\citeauthoryear{Duboue and McKeown}{Duboue and
  McKeown}{2003}]%
        {duboue2003statistical}
\bibfield{author}{\bibinfo{person}{Pablo~A Duboue} {and}
  \bibinfo{person}{Kathleen~R McKeown}.} \bibinfo{year}{2003}\natexlab{}.
\newblock \showarticletitle{Statistical acquisition of content selection rules
  for natural language generation}. In \bibinfo{booktitle}{\emph{Proceedings of
  the 2003 conference on Empirical methods in natural language processing}}.
  \bibinfo{pages}{121--128}.
\newblock


\bibitem[\protect\citeauthoryear{Durmus, He, and Diab}{Durmus
  et~al\mbox{.}}{2020}]%
        {durmus2020feqa}
\bibfield{author}{\bibinfo{person}{Esin Durmus}, \bibinfo{person}{He He}, {and}
  \bibinfo{person}{Mona Diab}.} \bibinfo{year}{2020}\natexlab{}.
\newblock \showarticletitle{FEQA: A Question Answering Evaluation Framework for
  Faithfulness Assessment in Abstractive Summarization}. In
  \bibinfo{booktitle}{\emph{Proceedings of the 58th Annual Meeting of the
  Association for Computational Linguistics}}. \bibinfo{pages}{5055--5070}.
\newblock


\bibitem[\protect\citeauthoryear{Ferreira, van~der Lee, Van~Miltenburg, and
  Krahmer}{Ferreira et~al\mbox{.}}{2019}]%
        {ferreira2019neural}
\bibfield{author}{\bibinfo{person}{Thiago~Castro Ferreira},
  \bibinfo{person}{Chris van~der Lee}, \bibinfo{person}{Emiel Van~Miltenburg},
  {and} \bibinfo{person}{Emiel Krahmer}.} \bibinfo{year}{2019}\natexlab{}.
\newblock \showarticletitle{Neural data-to-text generation: A comparison
  between pipeline and end-to-end architectures}. In
  \bibinfo{booktitle}{\emph{Proceedings of the 2019 Conference on Empirical
  Methods in Natural Language Processing and the 9th International Joint
  Conference on Natural Language Processing (EMNLP-IJCNLP)}}.
  \bibinfo{pages}{552--562}.
\newblock


\bibitem[\protect\citeauthoryear{Gao, Wu, Homayoun, and Zhao}{Gao
  et~al\mbox{.}}{2019}]%
        {gao2019dyngraph2seq}
\bibfield{author}{\bibinfo{person}{Yuyang Gao}, \bibinfo{person}{Lingfei Wu},
  \bibinfo{person}{Houman Homayoun}, {and} \bibinfo{person}{Liang Zhao}.}
  \bibinfo{year}{2019}\natexlab{}.
\newblock \showarticletitle{Dyngraph2seq: Dynamic-graph-to-sequence
  interpretable learning for health stage prediction in online health forums}.
  In \bibinfo{booktitle}{\emph{ICDM}}.
\newblock


\bibitem[\protect\citeauthoryear{Gardent, Shimorina, Narayan, and
  Perez-Beltrachini}{Gardent et~al\mbox{.}}{2017a}]%
        {gardent2017creating}
\bibfield{author}{\bibinfo{person}{Claire Gardent}, \bibinfo{person}{Anastasia
  Shimorina}, \bibinfo{person}{Shashi Narayan}, {and} \bibinfo{person}{Laura
  Perez-Beltrachini}.} \bibinfo{year}{2017}\natexlab{a}.
\newblock \showarticletitle{Creating Training Corpora for Micro-Planners}. In
  \bibinfo{booktitle}{\emph{Proceedings of the 55th Annual Meeting of the
  Association for Computational Linguistics}}. \bibinfo{address}{Vancouver,
  Canada}.
\newblock


\bibitem[\protect\citeauthoryear{Gardent, Shimorina, Narayan, and
  Perez-Beltrachini}{Gardent et~al\mbox{.}}{2017b}]%
        {gardent2017webnlg}
\bibfield{author}{\bibinfo{person}{Claire Gardent}, \bibinfo{person}{Anastasia
  Shimorina}, \bibinfo{person}{Shashi Narayan}, {and} \bibinfo{person}{Laura
  Perez-Beltrachini}.} \bibinfo{year}{2017}\natexlab{b}.
\newblock \showarticletitle{The webnlg challenge: Generating text from rdf
  data}. In \bibinfo{booktitle}{\emph{Proceedings of the 10th International
  Conference on Natural Language Generation}}. \bibinfo{pages}{124--133}.
\newblock


\bibitem[\protect\citeauthoryear{Gu, Lu, Li, and Li}{Gu et~al\mbox{.}}{2016}]%
        {gu2016incorporating}
\bibfield{author}{\bibinfo{person}{Jiatao Gu}, \bibinfo{person}{Zhengdong Lu},
  \bibinfo{person}{Hang Li}, {and} \bibinfo{person}{Victor~OK Li}.}
  \bibinfo{year}{2016}\natexlab{}.
\newblock \showarticletitle{Incorporating copying mechanism in
  sequence-to-sequence learning}.
\newblock \bibinfo{journal}{\emph{arXiv preprint arXiv:1603.06393}}
  (\bibinfo{year}{2016}).
\newblock


\bibitem[\protect\citeauthoryear{Hamilton, Ying, and Leskovec}{Hamilton
  et~al\mbox{.}}{2017}]%
        {hamilton2017inductive}
\bibfield{author}{\bibinfo{person}{William~L Hamilton}, \bibinfo{person}{Rex
  Ying}, {and} \bibinfo{person}{Jure Leskovec}.}
  \bibinfo{year}{2017}\natexlab{}.
\newblock \showarticletitle{Inductive representation learning on large graphs}.
  In \bibinfo{booktitle}{\emph{Proceedings of the 31st International Conference
  on Neural Information Processing Systems}}. \bibinfo{pages}{1025--1035}.
\newblock


\bibitem[\protect\citeauthoryear{Hao, Zhang, Liu, He, Liu, Wu, and Zhao}{Hao
  et~al\mbox{.}}{2017}]%
        {hao2017end}
\bibfield{author}{\bibinfo{person}{Yanchao Hao}, \bibinfo{person}{Yuanzhe
  Zhang}, \bibinfo{person}{Kang Liu}, \bibinfo{person}{Shizhu He},
  \bibinfo{person}{Zhanyi Liu}, \bibinfo{person}{Hua Wu}, {and}
  \bibinfo{person}{Jun Zhao}.} \bibinfo{year}{2017}\natexlab{}.
\newblock \showarticletitle{An end-to-end model for question answering over
  knowledge base with cross-attention combining global knowledge}. In
  \bibinfo{booktitle}{\emph{Proceedings of the 55th Annual Meeting of the
  Association for Computational Linguistics}}. \bibinfo{pages}{221--231}.
\newblock


\bibitem[\protect\citeauthoryear{Hochreiter and Schmidhuber}{Hochreiter and
  Schmidhuber}{1997}]%
        {hochreiter1997long}
\bibfield{author}{\bibinfo{person}{Sepp Hochreiter} {and}
  \bibinfo{person}{J{\"u}rgen Schmidhuber}.} \bibinfo{year}{1997}\natexlab{}.
\newblock \showarticletitle{Long short-term memory}.
\newblock \bibinfo{journal}{\emph{Neural computation}} \bibinfo{volume}{9},
  \bibinfo{number}{8} (\bibinfo{year}{1997}), \bibinfo{pages}{1735--1780}.
\newblock


\bibitem[\protect\citeauthoryear{Jagfeld, Jenne, and Vu}{Jagfeld
  et~al\mbox{.}}{2018}]%
        {jagfeld2018sequence}
\bibfield{author}{\bibinfo{person}{Glorianna Jagfeld}, \bibinfo{person}{Sabrina
  Jenne}, {and} \bibinfo{person}{Ngoc~Thang Vu}.}
  \bibinfo{year}{2018}\natexlab{}.
\newblock \showarticletitle{Sequence-to-Sequence Models for Data-to-Text
  Natural Language Generation: Word-vs. Character-based Processing and Output
  Diversity}.
\newblock \bibinfo{journal}{\emph{arXiv preprint arXiv:1810.04864}}
  (\bibinfo{year}{2018}).
\newblock


\bibitem[\protect\citeauthoryear{Kale and Rastogi}{Kale and Rastogi}{2020}]%
        {kale2020text}
\bibfield{author}{\bibinfo{person}{Mihir Kale} {and} \bibinfo{person}{Abhinav
  Rastogi}.} \bibinfo{year}{2020}\natexlab{}.
\newblock \showarticletitle{Text-to-Text Pre-Training for Data-to-Text Tasks}.
  In \bibinfo{booktitle}{\emph{Proceedings of the 13th International Conference
  on Natural Language Generation}}. \bibinfo{pages}{97--102}.
\newblock


\bibitem[\protect\citeauthoryear{Kingma and Ba}{Kingma and Ba}{2014}]%
        {kingma2014adam}
\bibfield{author}{\bibinfo{person}{Diederik~P Kingma} {and}
  \bibinfo{person}{Jimmy Ba}.} \bibinfo{year}{2014}\natexlab{}.
\newblock \showarticletitle{Adam: A method for stochastic optimization}.
\newblock \bibinfo{journal}{\emph{arXiv preprint arXiv:1412.6980}}
  (\bibinfo{year}{2014}).
\newblock


\bibitem[\protect\citeauthoryear{Kipf and Welling}{Kipf and Welling}{2016}]%
        {kipf2016semi}
\bibfield{author}{\bibinfo{person}{Thomas~N Kipf} {and} \bibinfo{person}{Max
  Welling}.} \bibinfo{year}{2016}\natexlab{}.
\newblock \showarticletitle{Semi-supervised classification with graph
  convolutional networks}.
\newblock \bibinfo{journal}{\emph{arXiv preprint arXiv:1609.02907}}
  (\bibinfo{year}{2016}).
\newblock


\bibitem[\protect\citeauthoryear{Krizhevsky, Sutskever, and Hinton}{Krizhevsky
  et~al\mbox{.}}{2012}]%
        {krizhevsky2012imagenet}
\bibfield{author}{\bibinfo{person}{Alex Krizhevsky}, \bibinfo{person}{Ilya
  Sutskever}, {and} \bibinfo{person}{Geoffrey~E Hinton}.}
  \bibinfo{year}{2012}\natexlab{}.
\newblock \showarticletitle{Imagenet classification with deep convolutional
  neural networks}.
\newblock \bibinfo{journal}{\emph{Advances in neural information processing
  systems}}  \bibinfo{volume}{25} (\bibinfo{year}{2012}),
  \bibinfo{pages}{1097--1105}.
\newblock


\bibitem[\protect\citeauthoryear{Li, Tarlow, Brockschmidt, and Zemel}{Li
  et~al\mbox{.}}{2015}]%
        {li2015gated}
\bibfield{author}{\bibinfo{person}{Yujia Li}, \bibinfo{person}{Daniel Tarlow},
  \bibinfo{person}{Marc Brockschmidt}, {and} \bibinfo{person}{Richard Zemel}.}
  \bibinfo{year}{2015}\natexlab{}.
\newblock \showarticletitle{Gated graph sequence neural networks}.
\newblock \bibinfo{journal}{\emph{arXiv preprint arXiv:1511.05493}}
  (\bibinfo{year}{2015}).
\newblock


\bibitem[\protect\citeauthoryear{Liu, Wang, Sha, Chang, and Sui}{Liu
  et~al\mbox{.}}{2018}]%
        {liu2018table}
\bibfield{author}{\bibinfo{person}{Tianyu Liu}, \bibinfo{person}{Kexiang Wang},
  \bibinfo{person}{Lei Sha}, \bibinfo{person}{Baobao Chang}, {and}
  \bibinfo{person}{Zhifang Sui}.} \bibinfo{year}{2018}\natexlab{}.
\newblock \showarticletitle{Table-to-text generation by structure-aware seq2seq
  learning}. In \bibinfo{booktitle}{\emph{Thirty-Second AAAI Conference on
  Artificial Intelligence}}.
\newblock


\bibitem[\protect\citeauthoryear{Luong, Pham, and Manning}{Luong
  et~al\mbox{.}}{2015}]%
        {luong2015effective}
\bibfield{author}{\bibinfo{person}{Minh-Thang Luong}, \bibinfo{person}{Hieu
  Pham}, {and} \bibinfo{person}{Christopher~D Manning}.}
  \bibinfo{year}{2015}\natexlab{}.
\newblock \showarticletitle{Effective approaches to attention-based neural
  machine translation}.
\newblock \bibinfo{journal}{\emph{arXiv preprint arXiv:1508.04025}}
  (\bibinfo{year}{2015}).
\newblock


\bibitem[\protect\citeauthoryear{Ma and Tang}{Ma and Tang}{2021}]%
        {ma2021deep}
\bibfield{author}{\bibinfo{person}{Yao Ma} {and} \bibinfo{person}{Jiliang
  Tang}.} \bibinfo{year}{2021}\natexlab{}.
\newblock \bibinfo{booktitle}{\emph{Deep Learning on Graphs}}.
\newblock \bibinfo{publisher}{Cambridge University Press}.
\newblock


\bibitem[\protect\citeauthoryear{Marcheggiani and
  Perez-Beltrachini}{Marcheggiani and Perez-Beltrachini}{2018}]%
        {marcheggiani2018deep}
\bibfield{author}{\bibinfo{person}{Diego Marcheggiani} {and}
  \bibinfo{person}{Laura Perez-Beltrachini}.} \bibinfo{year}{2018}\natexlab{}.
\newblock \showarticletitle{Deep Graph Convolutional Encoders for Structured
  Data to Text Generation}. In \bibinfo{booktitle}{\emph{Proceedings of the
  11th International Conference on Natural Language Generation}}.
  \bibinfo{pages}{1--9}.
\newblock


\bibitem[\protect\citeauthoryear{Marcheggiani and Titov}{Marcheggiani and
  Titov}{2017}]%
        {marcheggiani2017encoding}
\bibfield{author}{\bibinfo{person}{Diego Marcheggiani} {and}
  \bibinfo{person}{Ivan Titov}.} \bibinfo{year}{2017}\natexlab{}.
\newblock \showarticletitle{Encoding sentences with graph convolutional
  networks for semantic role labeling}.
\newblock \bibinfo{journal}{\emph{arXiv preprint arXiv:1703.04826}}
  (\bibinfo{year}{2017}).
\newblock


\bibitem[\protect\citeauthoryear{Miwa and Bansal}{Miwa and Bansal}{2016}]%
        {miwa2016end}
\bibfield{author}{\bibinfo{person}{Makoto Miwa} {and} \bibinfo{person}{Mohit
  Bansal}.} \bibinfo{year}{2016}\natexlab{}.
\newblock \showarticletitle{End-to-end relation extraction using lstms on
  sequences and tree structures}.
\newblock \bibinfo{journal}{\emph{arXiv preprint arXiv:1601.00770}}
  (\bibinfo{year}{2016}).
\newblock


\bibitem[\protect\citeauthoryear{Moryossef, Goldberg, and Dagan}{Moryossef
  et~al\mbox{.}}{2019}]%
        {moryossef2019step}
\bibfield{author}{\bibinfo{person}{Amit Moryossef}, \bibinfo{person}{Yoav
  Goldberg}, {and} \bibinfo{person}{Ido Dagan}.}
  \bibinfo{year}{2019}\natexlab{}.
\newblock \showarticletitle{Step-by-step: Separating planning from realization
  in neural data-to-text generation}.
\newblock \bibinfo{journal}{\emph{arXiv preprint arXiv:1904.03396}}
  (\bibinfo{year}{2019}).
\newblock


\bibitem[\protect\citeauthoryear{Nan, Radev, Zhang, Rau, Sivaprasad, Hsieh,
  Tang, Vyas, Verma, Krishna, Liu, Irwanto, Pan, Rahman, Zaidi, Mutuma,
  Tarabar, Gupta, Yu, Tan, Lin, Xiong, Socher, and Rajani}{Nan
  et~al\mbox{.}}{2021}]%
        {nan2021dart}
\bibfield{author}{\bibinfo{person}{Linyong Nan}, \bibinfo{person}{Dragomir
  Radev}, \bibinfo{person}{Rui Zhang}, \bibinfo{person}{Amrit Rau},
  \bibinfo{person}{Abhinand Sivaprasad}, \bibinfo{person}{Chiachun Hsieh},
  \bibinfo{person}{Xiangru Tang}, \bibinfo{person}{Aadit Vyas},
  \bibinfo{person}{Neha Verma}, \bibinfo{person}{Pranav Krishna},
  \bibinfo{person}{Yangxiaokang Liu}, \bibinfo{person}{Nadia Irwanto},
  \bibinfo{person}{Jessica Pan}, \bibinfo{person}{Faiaz Rahman},
  \bibinfo{person}{Ahmad Zaidi}, \bibinfo{person}{Murori Mutuma},
  \bibinfo{person}{Yasin Tarabar}, \bibinfo{person}{Ankit Gupta},
  \bibinfo{person}{Tao Yu}, \bibinfo{person}{Yi~Chern Tan},
  \bibinfo{person}{Xi~Victoria Lin}, \bibinfo{person}{Caiming Xiong},
  \bibinfo{person}{Richard Socher}, {and} \bibinfo{person}{Nazneen~Fatema
  Rajani}.} \bibinfo{year}{2021}\natexlab{}.
\newblock \showarticletitle{DART: Open-Domain Structured Data Record to Text
  Generation}.
\newblock \bibinfo{journal}{\emph{arXiv preprint arXiv:2007.02871}}
  (\bibinfo{year}{2021}).
\newblock


\bibitem[\protect\citeauthoryear{Nayak and Ng}{Nayak and Ng}{2020}]%
        {nayak2020effective}
\bibfield{author}{\bibinfo{person}{Tapas Nayak} {and} \bibinfo{person}{Hwee~Tou
  Ng}.} \bibinfo{year}{2020}\natexlab{}.
\newblock \showarticletitle{Effective modeling of encoder-decoder architecture
  for joint entity and relation extraction}. In
  \bibinfo{booktitle}{\emph{Proceedings of the AAAI Conference on Artificial
  Intelligence}}, Vol.~\bibinfo{volume}{34}. \bibinfo{pages}{8528--8535}.
\newblock


\bibitem[\protect\citeauthoryear{Papineni, Roukos, Ward, and Zhu}{Papineni
  et~al\mbox{.}}{2002}]%
        {papineni2002bleu}
\bibfield{author}{\bibinfo{person}{Kishore Papineni}, \bibinfo{person}{Salim
  Roukos}, \bibinfo{person}{Todd Ward}, {and} \bibinfo{person}{Wei-Jing Zhu}.}
  \bibinfo{year}{2002}\natexlab{}.
\newblock \showarticletitle{BLEU: a method for automatic evaluation of machine
  translation}. In \bibinfo{booktitle}{\emph{Proceedings of the 40th annual
  meeting on association for computational linguistics}}.
  \bibinfo{pages}{311--318}.
\newblock


\bibitem[\protect\citeauthoryear{Pouriyeh, Allahyari, Kochut, Cheng, and
  Arabnia}{Pouriyeh et~al\mbox{.}}{2017}]%
        {pouriyeh2017lda}
\bibfield{author}{\bibinfo{person}{Seyedamin Pouriyeh}, \bibinfo{person}{Mehdi
  Allahyari}, \bibinfo{person}{Krzysztof Kochut}, \bibinfo{person}{Gong Cheng},
  {and} \bibinfo{person}{Hamid~Reza Arabnia}.} \bibinfo{year}{2017}\natexlab{}.
\newblock \showarticletitle{Es-lda: Entity summarization using knowledge-based
  topic modeling}. In \bibinfo{booktitle}{\emph{Proceedings of the Eighth
  International Joint Conference on Natural Language Processing (Volume 1: Long
  Papers)}}. \bibinfo{pages}{316--325}.
\newblock


\bibitem[\protect\citeauthoryear{Raffel, Shazeer, Roberts, Lee, Narang, Matena,
  Zhou, Li, and Liu}{Raffel et~al\mbox{.}}{2019}]%
        {raffel2019exploring}
\bibfield{author}{\bibinfo{person}{Colin Raffel}, \bibinfo{person}{Noam
  Shazeer}, \bibinfo{person}{Adam Roberts}, \bibinfo{person}{Katherine Lee},
  \bibinfo{person}{Sharan Narang}, \bibinfo{person}{Michael Matena},
  \bibinfo{person}{Yanqi Zhou}, \bibinfo{person}{Wei Li}, {and}
  \bibinfo{person}{Peter~J Liu}.} \bibinfo{year}{2019}\natexlab{}.
\newblock \showarticletitle{Exploring the limits of transfer learning with a
  unified text-to-text transformer}.
\newblock \bibinfo{journal}{\emph{arXiv preprint arXiv:1910.10683}}
  (\bibinfo{year}{2019}).
\newblock


\bibitem[\protect\citeauthoryear{Rennie, Marcheret, Mroueh, Ross, and
  Goel}{Rennie et~al\mbox{.}}{2017}]%
        {rennie2017self}
\bibfield{author}{\bibinfo{person}{Steven~J Rennie}, \bibinfo{person}{Etienne
  Marcheret}, \bibinfo{person}{Youssef Mroueh}, \bibinfo{person}{Jerret Ross},
  {and} \bibinfo{person}{Vaibhava Goel}.} \bibinfo{year}{2017}\natexlab{}.
\newblock \showarticletitle{Self-critical sequence training for image
  captioning}. In \bibinfo{booktitle}{\emph{Proceedings of the IEEE Conference
  on Computer Vision and Pattern Recognition}}. \bibinfo{pages}{7008--7024}.
\newblock


\bibitem[\protect\citeauthoryear{Ribeiro, Zhang, Gardent, and Gurevych}{Ribeiro
  et~al\mbox{.}}{2020}]%
        {ribeiro2020modeling}
\bibfield{author}{\bibinfo{person}{Leonardo~FR Ribeiro}, \bibinfo{person}{Yue
  Zhang}, \bibinfo{person}{Claire Gardent}, {and} \bibinfo{person}{Iryna
  Gurevych}.} \bibinfo{year}{2020}\natexlab{}.
\newblock \showarticletitle{Modeling global and local node contexts for text
  generation from knowledge graphs}.
\newblock \bibinfo{journal}{\emph{Transactions of the Association for
  Computational Linguistics}}  \bibinfo{volume}{8} (\bibinfo{year}{2020}),
  \bibinfo{pages}{589--604}.
\newblock


\bibitem[\protect\citeauthoryear{Sahu, Christopoulou, Miwa, and Ananiadou}{Sahu
  et~al\mbox{.}}{2019}]%
        {sahu2019inter}
\bibfield{author}{\bibinfo{person}{Sunil~Kumar Sahu}, \bibinfo{person}{Fenia
  Christopoulou}, \bibinfo{person}{Makoto Miwa}, {and} \bibinfo{person}{Sophia
  Ananiadou}.} \bibinfo{year}{2019}\natexlab{}.
\newblock \showarticletitle{Inter-sentence Relation Extraction with
  Document-level Graph Convolutional Neural Network}. In
  \bibinfo{booktitle}{\emph{Proceedings of the 57th Annual Meeting of the
  Association for Computational Linguistics}}. \bibinfo{pages}{4309--4316}.
\newblock


\bibitem[\protect\citeauthoryear{See, Liu, and Manning}{See
  et~al\mbox{.}}{2017}]%
        {see2017get}
\bibfield{author}{\bibinfo{person}{Abigail See}, \bibinfo{person}{Peter~J Liu},
  {and} \bibinfo{person}{Christopher~D Manning}.}
  \bibinfo{year}{2017}\natexlab{}.
\newblock \showarticletitle{Get to the point: Summarization with
  pointer-generator networks}.
\newblock \bibinfo{journal}{\emph{arXiv preprint arXiv:1704.04368}}
  (\bibinfo{year}{2017}).
\newblock


\bibitem[\protect\citeauthoryear{Song, Gildea, Zhang, Wang, and Su}{Song
  et~al\mbox{.}}{2019}]%
        {song2019semantic}
\bibfield{author}{\bibinfo{person}{Linfeng Song}, \bibinfo{person}{Daniel
  Gildea}, \bibinfo{person}{Yue Zhang}, \bibinfo{person}{Zhiguo Wang}, {and}
  \bibinfo{person}{Jinsong Su}.} \bibinfo{year}{2019}\natexlab{}.
\newblock \showarticletitle{Semantic Neural Machine Translation using AMR}.
\newblock \bibinfo{journal}{\emph{arXiv preprint arXiv:1902.07282}}
  (\bibinfo{year}{2019}).
\newblock


\bibitem[\protect\citeauthoryear{Song, Zhang, Wang, and Gildea}{Song
  et~al\mbox{.}}{2018}]%
        {song2018graph}
\bibfield{author}{\bibinfo{person}{Linfeng Song}, \bibinfo{person}{Yue Zhang},
  \bibinfo{person}{Zhiguo Wang}, {and} \bibinfo{person}{Daniel Gildea}.}
  \bibinfo{year}{2018}\natexlab{}.
\newblock \showarticletitle{A graph-to-sequence model for AMR-to-text
  generation}.
\newblock \bibinfo{journal}{\emph{arXiv preprint arXiv:1805.02473}}
  (\bibinfo{year}{2018}).
\newblock


\bibitem[\protect\citeauthoryear{Stevens, Malone, Williams, Power, and
  Third}{Stevens et~al\mbox{.}}{2011}]%
        {stevens2011automating}
\bibfield{author}{\bibinfo{person}{Robert Stevens}, \bibinfo{person}{James
  Malone}, \bibinfo{person}{Sandra Williams}, \bibinfo{person}{Richard Power},
  {and} \bibinfo{person}{Allan Third}.} \bibinfo{year}{2011}\natexlab{}.
\newblock \showarticletitle{Automating generation of textual class definitions
  from OWL to English}. In \bibinfo{booktitle}{\emph{Journal of Biomedical
  Semantics}}, Vol.~\bibinfo{volume}{2}. BioMed Central, \bibinfo{pages}{S5}.
\newblock


\bibitem[\protect\citeauthoryear{Sui, Chen, Liu, Zhao, Zeng, and Liu}{Sui
  et~al\mbox{.}}{2020}]%
        {sui2020joint}
\bibfield{author}{\bibinfo{person}{Dianbo Sui}, \bibinfo{person}{Yubo Chen},
  \bibinfo{person}{Kang Liu}, \bibinfo{person}{Jun Zhao},
  \bibinfo{person}{Xiangrong Zeng}, {and} \bibinfo{person}{Shengping Liu}.}
  \bibinfo{year}{2020}\natexlab{}.
\newblock \showarticletitle{Joint entity and relation extraction with set
  prediction networks}.
\newblock \bibinfo{journal}{\emph{arXiv preprint arXiv:2011.01675}}
  (\bibinfo{year}{2020}).
\newblock


\bibitem[\protect\citeauthoryear{Trisedya, Qi, Zhang, and Wang}{Trisedya
  et~al\mbox{.}}{2018}]%
        {trisedya2018gtr}
\bibfield{author}{\bibinfo{person}{Bayu~Distiawan Trisedya},
  \bibinfo{person}{Jianzhong Qi}, \bibinfo{person}{Rui Zhang}, {and}
  \bibinfo{person}{Wei Wang}.} \bibinfo{year}{2018}\natexlab{}.
\newblock \showarticletitle{Gtr-lstm: A triple encoder for sentence generation
  from rdf data}. In \bibinfo{booktitle}{\emph{ACL}}, Vol.~\bibinfo{volume}{1}.
  \bibinfo{pages}{1627--1637}.
\newblock


\bibitem[\protect\citeauthoryear{Vaswani, Shazeer, Parmar, Uszkoreit, Jones,
  Gomez, Kaiser, and Polosukhin}{Vaswani et~al\mbox{.}}{2017}]%
        {vaswani2017attention}
\bibfield{author}{\bibinfo{person}{Ashish Vaswani}, \bibinfo{person}{Noam
  Shazeer}, \bibinfo{person}{Niki Parmar}, \bibinfo{person}{Jakob Uszkoreit},
  \bibinfo{person}{Llion Jones}, \bibinfo{person}{Aidan~N Gomez},
  \bibinfo{person}{{\L}ukasz Kaiser}, {and} \bibinfo{person}{Illia
  Polosukhin}.} \bibinfo{year}{2017}\natexlab{}.
\newblock \showarticletitle{Attention is all you need}. In
  \bibinfo{booktitle}{\emph{Advances in neural information processing
  systems}}. \bibinfo{pages}{5998--6008}.
\newblock


\bibitem[\protect\citeauthoryear{Veli{\v{c}}kovi{\'c}, Cucurull, Casanova,
  Romero, Lio, and Bengio}{Veli{\v{c}}kovi{\'c} et~al\mbox{.}}{2017}]%
        {velivckovic2017graph}
\bibfield{author}{\bibinfo{person}{Petar Veli{\v{c}}kovi{\'c}},
  \bibinfo{person}{Guillem Cucurull}, \bibinfo{person}{Arantxa Casanova},
  \bibinfo{person}{Adriana Romero}, \bibinfo{person}{Pietro Lio}, {and}
  \bibinfo{person}{Yoshua Bengio}.} \bibinfo{year}{2017}\natexlab{}.
\newblock \showarticletitle{Graph attention networks}.
\newblock \bibinfo{journal}{\emph{arXiv preprint arXiv:1710.10903}}
  (\bibinfo{year}{2017}).
\newblock


\bibitem[\protect\citeauthoryear{Wang, Yavuz, Lin, Ji, and Rajani}{Wang
  et~al\mbox{.}}{2021}]%
        {wang2021stage}
\bibfield{author}{\bibinfo{person}{Qingyun Wang}, \bibinfo{person}{Semih
  Yavuz}, \bibinfo{person}{Victoria Lin}, \bibinfo{person}{Heng Ji}, {and}
  \bibinfo{person}{Nazneen Rajani}.} \bibinfo{year}{2021}\natexlab{}.
\newblock \showarticletitle{Stage-wise Fine-tuning for Graph-to-Text
  Generation}.
\newblock \bibinfo{journal}{\emph{arXiv preprint arXiv:2105.08021}}
  (\bibinfo{year}{2021}).
\newblock


\bibitem[\protect\citeauthoryear{Wang, Yu, Zhang, Liu, Zhu, and Sun}{Wang
  et~al\mbox{.}}{2020}]%
        {wang2020tplinker}
\bibfield{author}{\bibinfo{person}{Yucheng Wang}, \bibinfo{person}{Bowen Yu},
  \bibinfo{person}{Yueyang Zhang}, \bibinfo{person}{Tingwen Liu},
  \bibinfo{person}{Hongsong Zhu}, {and} \bibinfo{person}{Limin Sun}.}
  \bibinfo{year}{2020}\natexlab{}.
\newblock \showarticletitle{TPLinker: Single-stage Joint Extraction of Entities
  and Relations Through Token Pair Linking}. In
  \bibinfo{booktitle}{\emph{Proceedings of the 28th International Conference on
  Computational Linguistics}}. \bibinfo{pages}{1572--1582}.
\newblock


\bibitem[\protect\citeauthoryear{Wu, Chen, Shen, Guo, Gao, Li, Pei, and
  Long}{Wu et~al\mbox{.}}{2021}]%
        {wu2021graph}
\bibfield{author}{\bibinfo{person}{Lingfei Wu}, \bibinfo{person}{Yu Chen},
  \bibinfo{person}{Kai Shen}, \bibinfo{person}{Xiaojie Guo},
  \bibinfo{person}{Hanning Gao}, \bibinfo{person}{Shucheng Li},
  \bibinfo{person}{Jian Pei}, {and} \bibinfo{person}{Bo Long}.}
  \bibinfo{year}{2021}\natexlab{}.
\newblock \showarticletitle{Graph Neural Networks for Natural Language
  Processing: A Survey}.
\newblock \bibinfo{journal}{\emph{arXiv preprint arXiv:2106.06090}}
  (\bibinfo{year}{2021}).
\newblock


\bibitem[\protect\citeauthoryear{Xu, Wu, Wang, Feng, Witbrock, and Sheinin}{Xu
  et~al\mbox{.}}{2018}]%
        {xu2018graph2seq}
\bibfield{author}{\bibinfo{person}{Kun Xu}, \bibinfo{person}{Lingfei Wu},
  \bibinfo{person}{Zhiguo Wang}, \bibinfo{person}{Yansong Feng},
  \bibinfo{person}{Michael Witbrock}, {and} \bibinfo{person}{Vadim Sheinin}.}
  \bibinfo{year}{2018}\natexlab{}.
\newblock \showarticletitle{Graph2seq: Graph to sequence learning with
  attention-based neural networks}.
\newblock \bibinfo{journal}{\emph{arXiv preprint arXiv:1804.00823}}
  (\bibinfo{year}{2018}).
\newblock


\bibitem[\protect\citeauthoryear{Xu, Mou, Li, Chen, Peng, and Jin}{Xu
  et~al\mbox{.}}{2015}]%
        {xu2015classifying}
\bibfield{author}{\bibinfo{person}{Yan Xu}, \bibinfo{person}{Lili Mou},
  \bibinfo{person}{Ge Li}, \bibinfo{person}{Yunchuan Chen},
  \bibinfo{person}{Hao Peng}, {and} \bibinfo{person}{Zhi Jin}.}
  \bibinfo{year}{2015}\natexlab{}.
\newblock \showarticletitle{Classifying relations via long short term memory
  networks along shortest dependency paths}. In
  \bibinfo{booktitle}{\emph{Proceedings of the 2015 conference on empirical
  methods in natural language processing}}. \bibinfo{pages}{1785--1794}.
\newblock


\bibitem[\protect\citeauthoryear{Yao, Wang, and Wan}{Yao et~al\mbox{.}}{2020}]%
        {yao2020heterogeneous}
\bibfield{author}{\bibinfo{person}{Shaowei Yao}, \bibinfo{person}{Tianming
  Wang}, {and} \bibinfo{person}{Xiaojun Wan}.} \bibinfo{year}{2020}\natexlab{}.
\newblock \showarticletitle{Heterogeneous graph transformer for
  graph-to-sequence learning}. In \bibinfo{booktitle}{\emph{Proceedings of the
  58th Annual Meeting of the Association for Computational Linguistics}}.
  \bibinfo{pages}{7145--7154}.
\newblock


\bibitem[\protect\citeauthoryear{Zelenko, Aone, and Richardella}{Zelenko
  et~al\mbox{.}}{2003}]%
        {zelenko2003kernel}
\bibfield{author}{\bibinfo{person}{Dmitry Zelenko}, \bibinfo{person}{Chinatsu
  Aone}, {and} \bibinfo{person}{Anthony Richardella}.}
  \bibinfo{year}{2003}\natexlab{}.
\newblock \showarticletitle{Kernel methods for relation extraction}.
\newblock \bibinfo{journal}{\emph{Journal of machine learning research}}
  \bibinfo{volume}{3}, \bibinfo{number}{Feb} (\bibinfo{year}{2003}),
  \bibinfo{pages}{1083--1106}.
\newblock


\bibitem[\protect\citeauthoryear{Zeng, Liu, Lai, Zhou, and Zhao}{Zeng
  et~al\mbox{.}}{2014}]%
        {zeng2014relation}
\bibfield{author}{\bibinfo{person}{Daojian Zeng}, \bibinfo{person}{Kang Liu},
  \bibinfo{person}{Siwei Lai}, \bibinfo{person}{Guangyou Zhou}, {and}
  \bibinfo{person}{Jun Zhao}.} \bibinfo{year}{2014}\natexlab{}.
\newblock \showarticletitle{Relation classification via convolutional deep
  neural network}. In \bibinfo{booktitle}{\emph{Proceedings of COLING 2014, the
  25th International Conference on Computational Linguistics: Technical
  Papers}}. \bibinfo{pages}{2335--2344}.
\newblock


\bibitem[\protect\citeauthoryear{Zeng, Zhang, and Liu}{Zeng
  et~al\mbox{.}}{2020}]%
        {zeng2020copymtl}
\bibfield{author}{\bibinfo{person}{Daojian Zeng}, \bibinfo{person}{Haoran
  Zhang}, {and} \bibinfo{person}{Qianying Liu}.}
  \bibinfo{year}{2020}\natexlab{}.
\newblock \showarticletitle{Copymtl: Copy mechanism for joint extraction of
  entities and relations with multi-task learning}. In
  \bibinfo{booktitle}{\emph{Proceedings of the AAAI Conference on Artificial
  Intelligence}}, Vol.~\bibinfo{volume}{34}. \bibinfo{pages}{9507--9514}.
\newblock


\bibitem[\protect\citeauthoryear{Zeng, Zeng, He, Liu, and Zhao}{Zeng
  et~al\mbox{.}}{2018}]%
        {zeng2018extracting}
\bibfield{author}{\bibinfo{person}{Xiangrong Zeng}, \bibinfo{person}{Daojian
  Zeng}, \bibinfo{person}{Shizhu He}, \bibinfo{person}{Kang Liu}, {and}
  \bibinfo{person}{Jun Zhao}.} \bibinfo{year}{2018}\natexlab{}.
\newblock \showarticletitle{Extracting relational facts by an end-to-end neural
  model with copy mechanism}. In \bibinfo{booktitle}{\emph{Proceedings of the
  56th Annual Meeting of the Association for Computational Linguistics (Volume
  1: Long Papers)}}. \bibinfo{pages}{506--514}.
\newblock


\bibitem[\protect\citeauthoryear{Zhang, Zhang, and Fu}{Zhang
  et~al\mbox{.}}{2017}]%
        {zhang2017end}
\bibfield{author}{\bibinfo{person}{Meishan Zhang}, \bibinfo{person}{Yue Zhang},
  {and} \bibinfo{person}{Guohong Fu}.} \bibinfo{year}{2017}\natexlab{}.
\newblock \showarticletitle{End-to-end neural relation extraction with global
  optimization}. In \bibinfo{booktitle}{\emph{Proceedings of the 2017
  Conference on Empirical Methods in Natural Language Processing}}.
  \bibinfo{pages}{1730--1740}.
\newblock


\bibitem[\protect\citeauthoryear{Zhao, Xu, Cheng, Li, and Gao}{Zhao
  et~al\mbox{.}}{2021}]%
        {zhao2021representation}
\bibfield{author}{\bibinfo{person}{Kang Zhao}, \bibinfo{person}{Hua Xu},
  \bibinfo{person}{Yue Cheng}, \bibinfo{person}{Xiaoteng Li}, {and}
  \bibinfo{person}{Kai Gao}.} \bibinfo{year}{2021}\natexlab{}.
\newblock \showarticletitle{Representation iterative fusion based on
  heterogeneous graph neural network for joint entity and relation extraction}.
\newblock \bibinfo{journal}{\emph{Knowledge-Based Systems}}
  \bibinfo{volume}{219} (\bibinfo{year}{2021}), \bibinfo{pages}{106888}.
\newblock


\bibitem[\protect\citeauthoryear{Zheng, Wang, Bao, Hao, Zhou, and Xu}{Zheng
  et~al\mbox{.}}{2017}]%
        {zheng2017joint}
\bibfield{author}{\bibinfo{person}{Suncong Zheng}, \bibinfo{person}{Feng Wang},
  \bibinfo{person}{Hongyun Bao}, \bibinfo{person}{Yuexing Hao},
  \bibinfo{person}{Peng Zhou}, {and} \bibinfo{person}{Bo Xu}.}
  \bibinfo{year}{2017}\natexlab{}.
\newblock \showarticletitle{Joint extraction of entities and relations based on
  a novel tagging scheme}.
\newblock \bibinfo{journal}{\emph{arXiv preprint arXiv:1706.05075}}
  (\bibinfo{year}{2017}).
\newblock


\bibitem[\protect\citeauthoryear{Zhu, Wan, Zhou, Chen, Qiu, Zhang, Jiang, and
  Yu}{Zhu et~al\mbox{.}}{2019}]%
        {zhu2019triple}
\bibfield{author}{\bibinfo{person}{Yaoming Zhu}, \bibinfo{person}{Juncheng
  Wan}, \bibinfo{person}{Zhiming Zhou}, \bibinfo{person}{Liheng Chen},
  \bibinfo{person}{Lin Qiu}, \bibinfo{person}{Weinan Zhang},
  \bibinfo{person}{Xin Jiang}, {and} \bibinfo{person}{Yong Yu}.}
  \bibinfo{year}{2019}\natexlab{}.
\newblock \showarticletitle{Triple-to-Text: Converting RDF Triples into
  High-Quality Natural Languages via Optimizing an Inverse KL Divergence}.
\newblock \bibinfo{journal}{\emph{arXiv preprint arXiv:1906.01965}}
  (\bibinfo{year}{2019}).
\newblock


\end{thebibliography}

\appendix









\end{document}